\documentclass[sn-basic,Numbered]{sn-jnl}
\usepackage{amsmath,amsfonts}

\usepackage{graphicx}%
\usepackage{multirow}%
\usepackage{amsthm}%
\usepackage{mathrsfs}%
\usepackage[title]{appendix}%
\usepackage{textcomp}%
\usepackage{manyfoot}%
\usepackage{booktabs}%
\usepackage{algorithmicx}%
\usepackage{algpseudocode}%
\usepackage{listings}%

\usepackage[table]{xcolor}
\usepackage{algorithm}
\usepackage{array}
\usepackage{booktabs}
\usepackage{arydshln}
\usepackage{textcomp}
\usepackage{stfloats}
\usepackage{url}
\usepackage{verbatim}
\usepackage{graphicx}
\usepackage{color}
\usepackage{multirow}
\usepackage{pgfplots}
\usepackage{overpic}
\usepackage{subfigure}
\usepackage{amsfonts,amssymb}
\usepackage{bm}
\usepackage{makecell} 
\usepackage{hhline}
\usepackage{nicematrix}
\usepackage{indentfirst}

\definecolor{light-gray}{gray}{0.5}

\usepackage{hyperref}

\def\ie{\emph{i.e.}}
\def\eg{\emph{e.g.}}

\def\etal{{\em et al.~}}
\newcommand{\sota}{state-of-the-art~}
\usepackage{colortbl}
\usepackage{cleveref}
\crefformat{section}{\S#2#1#3} 
\crefformat{subsection}{\S#2#1#3}
\crefformat{subsubsection}{\S#2#1#3}

\begin{document}

\title[Article Title]{Lightweight Improved Residual Network for Efficient Inverse Tone Mapping}

\author[1]{\fnm{Liqi} \sur{Xue}}
\author[1]{\fnm{Tianyi} \sur{Xu}}
\author[2]{\fnm{Yongbao} \sur{Song}}
\author*[1]{\fnm{Yan} \sur{Liu}}\email{liuyan23@nankai.edu.cn}
\author[3]{\fnm{Lei} \sur{Zhang}}
\author[3]{\fnm{Xiantong} \sur{Zhen}}
\author*[1,4]{\fnm{Jun} \sur{Xu}}\email{nankaimathxujun@gmail.com}

\affil[1]{\orgdiv{School of Statistics and Data Science}, \orgname{Nankai University}, \orgaddress{\city{Tianjin}, \postcode{300071}, \country{China}}}
\affil[2]{\orgdiv{School of Mathematical Science}, \orgname{Nankai University}, \orgaddress{\city{Tianjin}, \postcode{300071}, \country{China}}}
\affil[3]{\orgdiv{Computer Science College}, \orgname{Guangdong University of Petrochemical Technology}, \orgaddress{\city{Maoming}, \postcode{525000}, \country{China}}}
\affil[4]{\orgdiv{Guangdong Provincial
Key Laboratory of Big Data Computing}, \orgname{The Chinese University of Hong
Kong (Shenzhen)}, \orgaddress{\city{Shenzhen}, \postcode{518172}, \country{China}}}


\abstract{

The display devices like HDR10 televisions are increasingly prevalent in our daily life for visualizing high dynamic range (HDR) images. But the majority of media images on the internet remain in 8-bit standard dynamic range (SDR) format. Therefore, converting SDR images to HDR ones by inverse tone mapping (ITM) is crucial to unlock the full potential of abundant media images. However, existing ITM methods are usually developed with complex network architectures requiring huge computational costs. In this paper, we propose a lightweight Improved Residual Network (IRNet) by enhancing the power of popular residual block for efficient ITM. Specifically, we propose a new Improved Residual Block (IRB) to extract and fuse multi-layer features for fine-grained HDR image reconstruction. Experiments on three benchmark datasets demonstrate that our IRNet achieves state-of-the-art performance on both the ITM and joint SR-ITM tasks. The code, models and data will be publicly available at \url{https://github.com/ThisisVikki/ITM-baseline}.

}


\maketitle

\section{Introduction}
High dynamic range (HDR) images defined in Rec.2020~\cite{rec.2020} exhibit clearer details in highlights and shadows, as well as smoother transitions on brightness and color, than the standard dynamic range (SDR) images with 8-bit color depth defined in Rec.709~\cite{rec.709}. Owing to these benefits, the manufacturers of television and mobile devices make a push to bring HDR contents to demanding consumers. Though HDR display devices allow more visually-pleasing contents in HDR images by Dolby Vision, HDR10, and HDR10+ technologies~\cite{Yao2023UHD}, the SDR images in 8 bit-depth would be featureless when being directly broadcast on HDR display devices~\cite{9008274}. To present the SDR images closer to human perception on HDR display devices, it is essential to convert SDR images into comfortable HDR ones without color or information loss. This challenging problem is known as inverse tone mapping (ITM)~\cite{9008274}, which has been studied in a more general sense rather than expanding the luminance range of camera raw image files in linear color space~\cite{chengHDRTVRe2022}.


Early image ITM methods mainly resort to global or local image processing operators for promising performance. Global ITM operators~\cite{Aky2007,Masia2009,Masia2017,6915289} usually utilize reverse tone mapping functions to extend the dynamic range of image pixels. But this would bring distorted details and uneven transitions between neighborhood pixels in different levels of brightness. Local ITM operators~\cite{2007Meylan,Banterle08} expand the image bit depth in a spatially-varying manner. Unfortunately, these methods would fail to preserve the global consistency of luminance ranges across an image. Recently, deep neural networks have been employed to tackle the ITM task from a data-driven perspective~\cite{endo,Marnerides2018Expandnet,santos2020single,liu2020single}. These networks usually contain strong backbones with complex architectures, which may require huge computational costs for promising ITM performance. Besides, the methods of~\cite{9008274,gan,Kim_Oh_Kim_2020} simultaneously tackle the joint image super-resolution (SR) and ITM (joint SR-ITM) tasks by separating the base and detail components from the input image with extra image decomposition~\cite{he2013guided}. However, this would further increase the model complexity and computational costs of current joint SR-ITM methods over previous ITM ones.

Despite their promising performance, the above-mentioned ITM methods suffer from two main limitations. Firstly, the complex model architectures obscure the core of the ITM problem, that is, ``expanding the luminance range of a low dynamic range image to produce a higher dynamic range image''~\cite{banterle2006inverse}. This problem of extending luminance range or color bit depth is similar to the tasks of image super-resolution~\cite{Hou2023FaceSR,Hu2022SR} and video frame prediction~\cite{xu2021temporal,hu2023dmvfn}, all of which aim to increase the highly-correlated information of the input image at different aspects. Therefore, it is possible to tackle the ITM problem by simple and lightweight neural networks, as inspired by the concurrent works in image super-resolution~\cite{classsr,Hu2022SR} and video frame prediction~\cite{hu2023dmvfn}. Secondly, the huge computational costs also limits the prevalence of ITM methods from being deployed into edge devices. For example, to perform ITM on a 4K-resolution (3840$\times$2160) image, Deep SR-ITM~\cite{9008274} needs 2.50M parameters, $\sim$1.05$\times10^4$G FLOPs, and a speed of 777.95ms, while HDRTVNet~\cite{chen2021new} needs 37.20M parameters, $\sim$1.41$\times10^4$G FLOPs, and a speed of 1513.43ms.

\begin{figure}
\centering
\begin{overpic}[width=12cm]{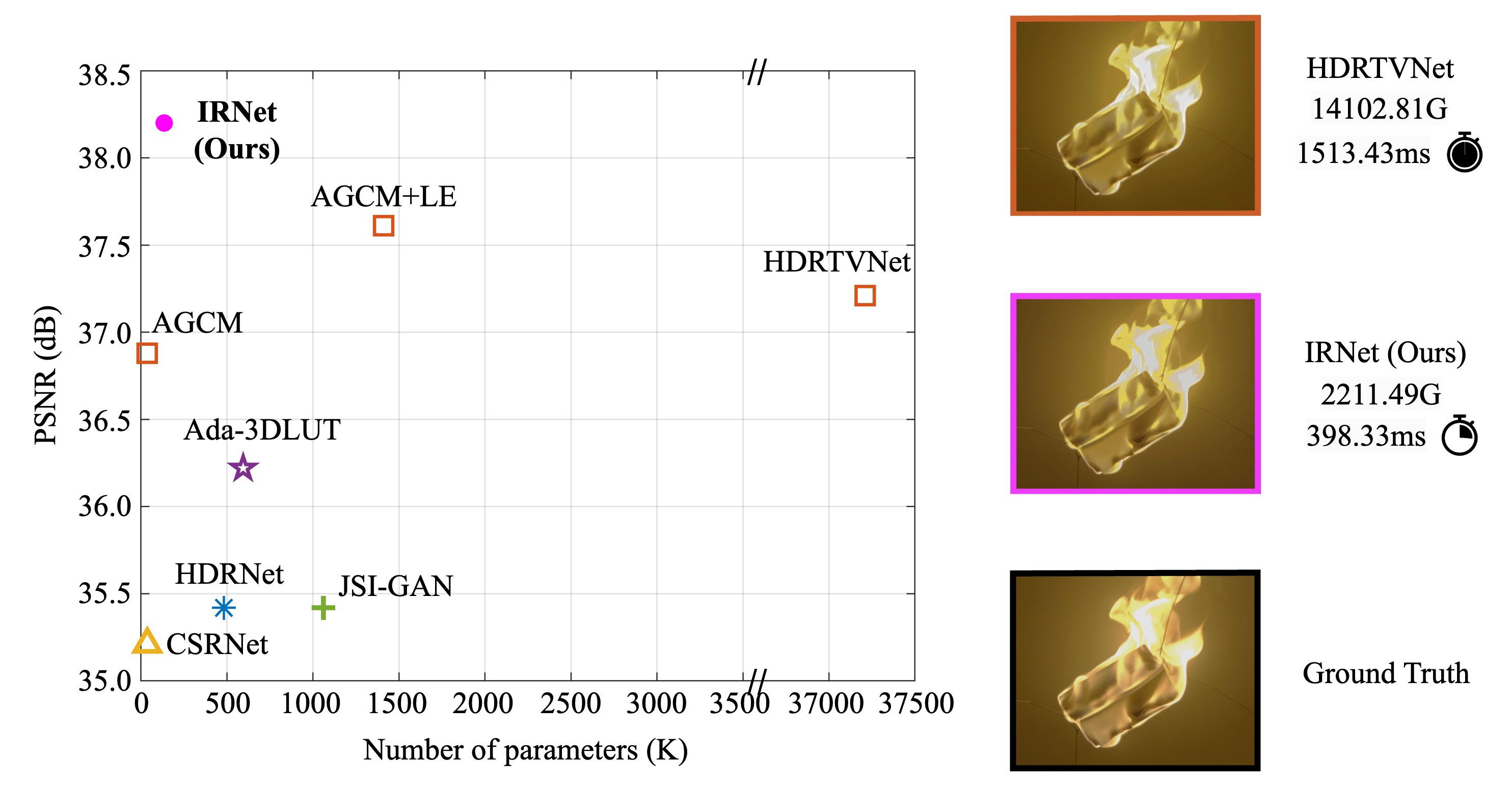}
\end{overpic}
\caption{\textbf{Parameter amounts and PSNR results by different ITM methods} on the test set of the HDRTV1K dataset~\cite{chen2021new}.
The visual quality, FLOPs and running time (visualized on the clock) on an SDR image of size ${3840 \times 2160 \times 3}$ are also provided for reference. The running time of HDRTVNet \cite{chen2021new} and our IRNet are 1513.43ms and 398.33ms, respectively.
}
\label{fig:compare}
\vspace{-7mm}
\end{figure}

In this paper, we leverage the popular residual learning recipe~\cite{he2016deep} to develop a simple and lightweight Improved Residual Network (IRNet) for efficient ITM performance. Specifically, we propose an Improved Residual Block (IRB) with simple modifications of the residual block~\cite{he2016deep} for fine-grained feature extraction and fusion. On network design, we also adopt the plain residual learning framework to avoid complex multi-branch architecture~\cite{9008274,Kim_Oh_Kim_2020}. Experiments on three benchmark datasets, including our newly collected one, show that our IRNet is very efficient and outperforms \sota ITM methods. As shown in Figure~\ref{fig:compare}, our IRNet only needs $\sim$0.13M parameters and $\sim$0.22$\times10^4$G FLOPs at a speed of 398.33ms to process a 4K-resolution image, which outperforms state-of-the-art methods on the ITM task. On the HDRTV1K dataset~\cite{chen2021new}, our IRNet exceeds AGCM+LE~\cite{chen2021new} on visual quality and PSNR by 0.59dB, but only has one tenth of parameter amounts ($\sim$0.13M \textsl{v.s.} $\sim$1.41M). Besides, our IRNet also achieves superior performance to Deep SR-ITM~\cite{9008274} and JSI-GAN~\cite{Kim_Oh_Kim_2020} on joint SR-ITM.

In summary, our main contributions are three-fold:
\begin{itemize}
\item We develop a lightweight Improved Residual Network (IRNet) for efficient image inverse tone mapping (ITM). Our IRNet is built upon a new Improved Residual Block (IRB) customized from the popular residual block for fine-grained feature extraction and fusion.
\item We collect a new test set for ITM, \ie, ITM-4K, that has 160 4K-resolution images of versatile scenes with ground truth HDR images. It serves as a good supplementary to HDRTV1K~\cite{chen2021new} that has 117 test images.
\item Experiments on the HDRTV1K dataset~\cite{chen2021new}, our new ITM-4K test set, and the test set in \cite{9008274} show that our lightweight IRNet is efficient and achieves impressive quantitative and qualitative results on the ITM and joint SR-ITM tasks. Comprehensive ablation studies also validate the effectiveness of our model design.

\end{itemize}


The rest of this paper is organized as follows. In \S\ref{related work}, we summarize the related work. In \S\ref{method}, we present the proposed Improved Residual Network (IRNet). In \S\ref{sec:experiments}, we perform experiments to validate the efficiency of our IRNet on ITM and joint SR-ITM. In \S\ref{conclusion}, we conclude this paper.

\section{Related Work \label{related work}}
\subsection{Inverse Tone Mapping}
The inverse tone mapping (ITM) task aims to transform a standard dynamic range (SDR, usually in 8-bit) image into a high dynamic range (HDR, usually in 16-bit) image. This problem is ill-posed due to the information loss in the luminance ranges of SDR images. Early explorations on the ITM task can be divided into global and local ITM operators. While the global ITM operators equally apply linear expansion~\cite{Aky2007}, cross-bilateral filtering \cite{6915289}, or a gamma-based expansion \cite{Masia2009,Masia2017} to all the pixels or patches of an input SDR image, the local ITM operators~\cite{2007Meylan,Banterle08} reconstruct highlight regions or expand the luminance ranges of each pixel or patch according to the local information around it. Previous works show that global ITM operators~\cite{Aky2007,6915289,Masia2009,Masia2017} could avoid undesired artifacts, but result in rough details and unnatural transitions due to the ignorance of local detail reconstruction. On the contrary, local ITM operators~\cite{2007Meylan,Banterle08} implemented adaptively on small areas would fail to capture the global consistency of luminance ranges.

To deal with the issues of locally undesired artifacts and global luminance consistency raised by the early methods mentioned above, many recent ITM methods~\cite{endo,eilertsen2017hdr,Marnerides2018Expandnet,santos2020single,9447972} shift to utilize the advancements of deep convolutional neural networks (CNNs). 
Early CNN-based methods~\cite{endo,9447972} merge low dynamic range (LDR) images captured under multiple exposure settings to produce an SDR image. Meanwhile, the work of~\cite{Marnerides2018Expandnet} presents a multi-branch CNN to implement ITM both on global and local perspectives. Then, the method of~\cite{santos2020single} introduces a feature masking strategy to address the problem of undesired artifacts emerged during the image reconstruction. Recently, the physical principle of HDR image formation is also incorporated into the designing of ITM CNNs~\cite{liu2020single,chen2021new}. For example, HDRTVNet~\cite{chen2021new} contains of an adaptive global color mapping network, a local enhancement network, and a highlight generation network.
Despite their promising performance, most of these methods require huge parameter amounts and computational costs, which hinders them from being deployed into resource-constrained edge devices. In this paper, we aim to develop a lightweight yet efficient ITM network.

\subsection{Joint Super-Resolution and Inverse Tone Mapping}
Joint Super-Resolution and Inverse Tone Mapping (joint SR-ITM) aims to simultaneously increase the spatial resolution and dynamic range of an input low-resolution and standard dynamic range (LR-SDR) image. Deep convolutional neural networks have also been applied to tackle the joint SR-ITM task~\cite{9008274,Kim_Oh_Kim_2020,Lecouat2022,Tan2023}. Considering that the luminance ranges of different image areas should be expanded adaptively, the method of~\cite{9008274} firstly decomposes an SDR image into a low-frequency structure component and a high-frequency detail component, and then processes the two components by two different but correlated network branches. The separation is implemented by guided-filtering~\cite{he2013guided}, which is widely used in image smoothing~\cite{Xu2021Smoothing}. This framework is also employed in the subsequent work of JSI-GAN~\cite{Kim_Oh_Kim_2020}. To tackle multi-frame SDR inputs, Lecouat \etal \cite{Lecouat2022} reformulated the joint SR-ITM task as an optimization problem to fuse multiple LR-SDR raw image bursts in different exposures into an HR-HDR image. Tan \etal \cite{Tan2023} developed a two-branch network to fuse a series of LR-LDR dynamic images into an HR-HDR one by estimating the motion cues~\cite{xu2021temporal}.

Though with appealing performance, the image decomposition based methods usually require multi-branch network architectures for the joint SR-ITM task, which, however, usually implies a considerable growth of parameter amounts and computational burden to tackle parallel feature extraction and elaborate interaction. In this paper, we propose a lightweight ITM network for inference efficiency, inspired by the merits of lightweight image super-resolution networks~\cite{dong2016accelerating,Liu2020}. 
\subsection{Efficient Image Processing}
Efficient image processing encompasses three crucial objectives: computational efficiency, memory efficiency, and effective feature utilization ~\cite{Hu2022SR,hu2023dmvfn,liang2023clusterformer,2023_transformer_flow,wang-etal-2023-mustie,2021_sg_net,Li_2018_ECCV}.
In pursuit of these objectives, extensive prior research has explored a range of techniques. 
Many prior works, for instance, utilize the Laplacian pyramid~\cite{burt1987laplacian,afifi2021learning}, which decomposes the input image into a low-resolution base layer, consuming the majority of computations, and several high-resolution detail layers that require significantly fewer computations~\cite{liang2021high}. 
Bilateral grid learning~\cite{chen2016,gharbi2017deep} is also a popular solution. This method learns approximate operators on downsampled images and subsequently executes these learned operators on the original images. Recently, due to the powerful representation capability of  transformers~\cite{2021mobieVit,2022mobileVit2,2023_transformer_flow, wang-etal-2023-mustie,liang2023clusterformer}, transformed-based architectures have been widely employed to augment feature utilization. While these methods have yielded promising results, most of them involve complicated architectures or sophistical pipelines. By contrast, residual learning~\cite{he2016deep} is a cheap yet effective method to enhance feature utilization by adding a skip connection to aggregate feature maps of different layers. This method has witnessed great success  in various computer vision tasks, such as super-resolution~\cite{Liu2020}, visual localization~\cite{9414517,liu2021densernet}, object detection~\cite{cui2021tf} and image classification~\cite{res2net}.

Several light-weight models~\cite{2021_sg_net,liu2021densernet} also use the residual structure, but mainly focus on multi-scale feature aggregation across blocks. For example,~\cite{liu2021densernet} adds feature extraction branches paralleled with the backbone to obtain multi-scale features from different blocks. Although this design is efficient due to the shared features and parameters along the main branch, the feature extraction branches also incur extra parameters and the fine-grained multi-scale feature fusion within each block is overlooked. Res2Net~\cite{res2net} explores multi-scale feature fusion within block by splitting feature maps along the channel dimension and transforming them using a set of group filters with progressive feature summation. RFDN~\cite{Liu2020} also considers feature fusion within block, however, it suffers from multiple information distillation steps, resulting in more resource consumption. In this paper, we also explore multi-scale feature fusion within our proposed Improved Residual Block (IRB) for image ITM. Our IRNet can well process a 4K-resolution SDR image in 0.4 seconds with $\sim$134K parameters.
\section{Proposed Method}
\label{method}

\subsection{Motivation}
\label{sec:motivation}

\begin{figure*}
    \centering
    \includegraphics[width=13cm]{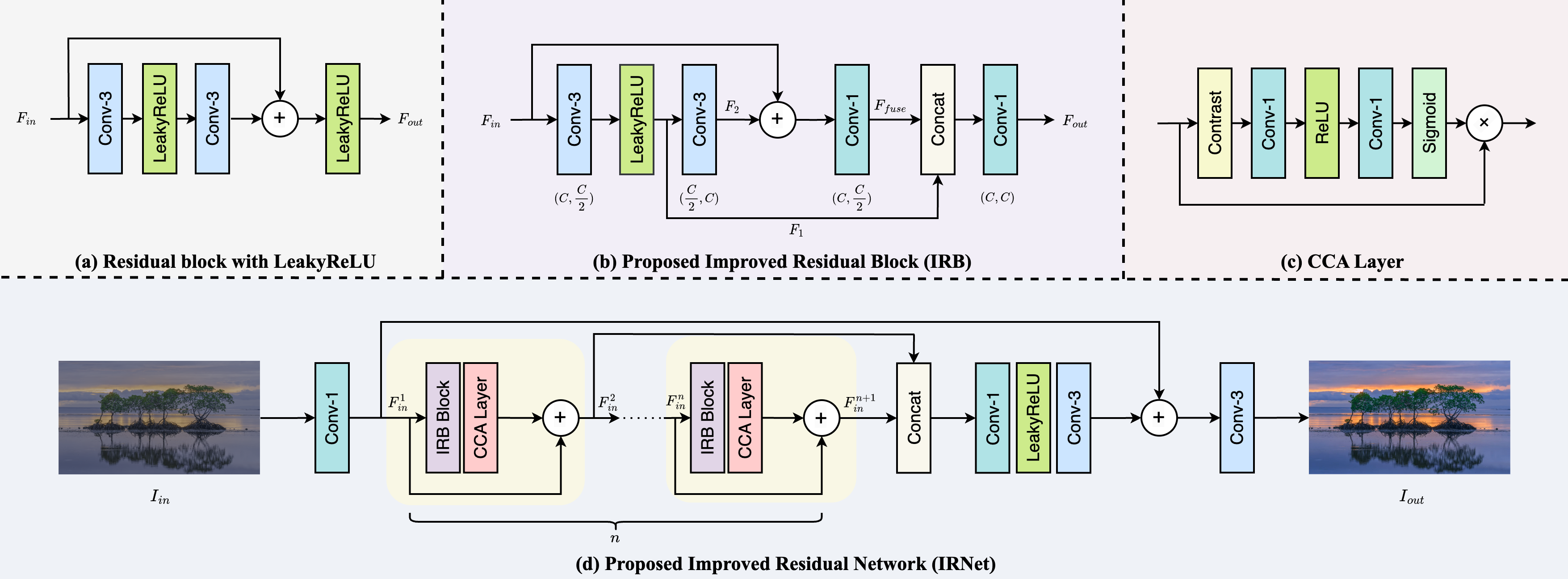}
\centering
\vspace{-2mm}
\caption{ {\bfseries Architectures of our Improved Residual Network (IRNet)}.
(a) The original residual block with LeakyReLU~\cite{he2016deep}.
(b) The proposed Improved Residual Block (IRB).
(c) The Contrast-aware Channel Attention layer (CCA)~\cite{CCA}.
(d) The proposed improved residual network (IRNet) for the ITM and joint SR-ITM tasks, which contains $n$ groups of IRB block and CCA layer~\cite{CCA} . We set $n=2$ for the ITM task and $n=5$ for the joint SR-ITM task, respectively.
}
\label{fig:arch}
\vspace{-3mm}
\end{figure*}

In the scene-referred workflow~\cite{chen2021new}, an HDR raw image in camera color space (usually in 16-bit color depth) will be tone mapped to an SDR RGB image in display-referred color space (usually in 8-bit color depth). This process is usually implemented in a camera imaging pipeline containing multiple image processing operations, during which different pixels usually undergo different compression strengths on dynamic ranges to produce visually pleasing image contrasts~\cite{Liang_2018_CVPR}.

\begin{figure*}
\centering
\includegraphics[width=13cm]{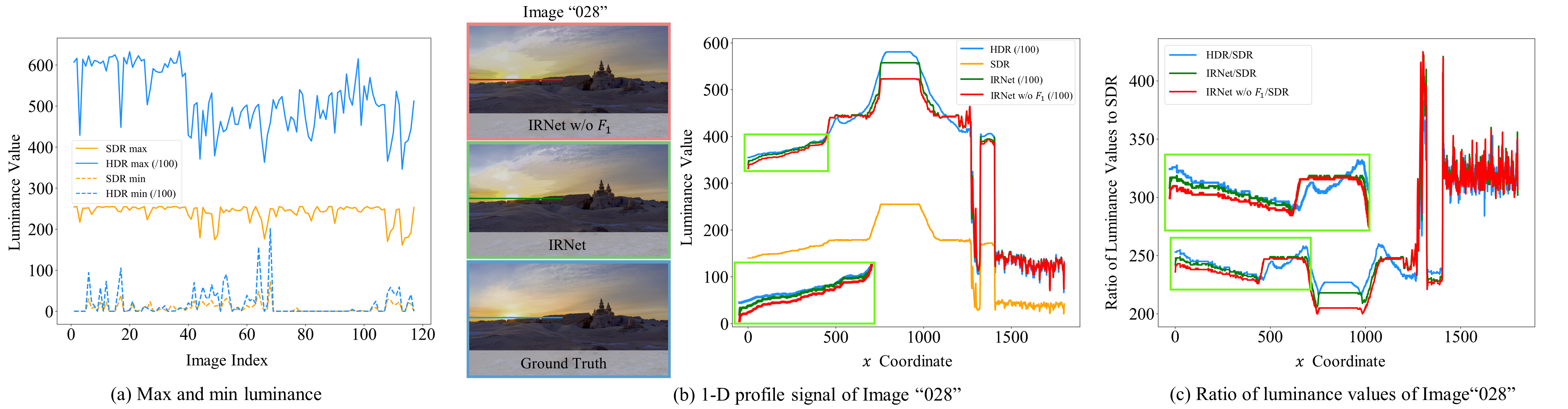}
\vspace{-2mm}
\caption{\textbf{Visualization of the luminance values on the HDRTV1K \cite{chen2021new} test set}. ``/100'' means that the values are divided by 100 to be better plotted.
(a) The maximum and minimum luminance values of the HDR images and the luminance values of the SDR images at the corresponding positions.
(b) Visual comparison of outputs by our IRNet with or w/o $\bm{F_1}$ on Image ``028''. The pixel values of 1-D profile signal in each image are plotted along the $x$ coordinate.
(c) The ratio of luminance values in the 1-D line pixels by our IRNet and IRNet w/o $\bm{F_1}$ to those from the corresponding SDR image.
}
\label{fig:luminance&1-D}
\centering
\vspace{-3mm}
\end{figure*}

The task of inverse tone mapping (ITM) aims to increase the dynamic range of light intensity (or luminance) in an SDR image. An SDR image in 8-bit depth can display a maximum of around 16.7 million shades of color, while an HDR image in 10-bit depth can display a maximum of around 1.07 billion shades of color, allowing it to exhibit more colors with better visual quality~\cite{Boitard2018ColorGamut}. To better understand the luminance difference between SDR and HDR images, in Figure~\ref{fig:luminance&1-D} (a), we visualize the maximum and minimum luminance values of 117 HDR images from the test set of~\cite{chen2021new}, as well as the luminance values at the corresponding positions of the paired SDR images. We observe that there are obvious gaps between the maximum values of HDR images and the values at the corresponding positions of the paired SDR images, whilst 
slight differences between the minimum values of the HDR images and the values at the corresponding positions of the paired SDR images. This indicates that high-luminance values change more greatly than low-luminance ones. Besides, the luminance values of different HDR images also show distinct gaps when compared to those values at the corresponding positions in the paired SDR images.

\begin{figure*}
\centering
\includegraphics[width=13cm]{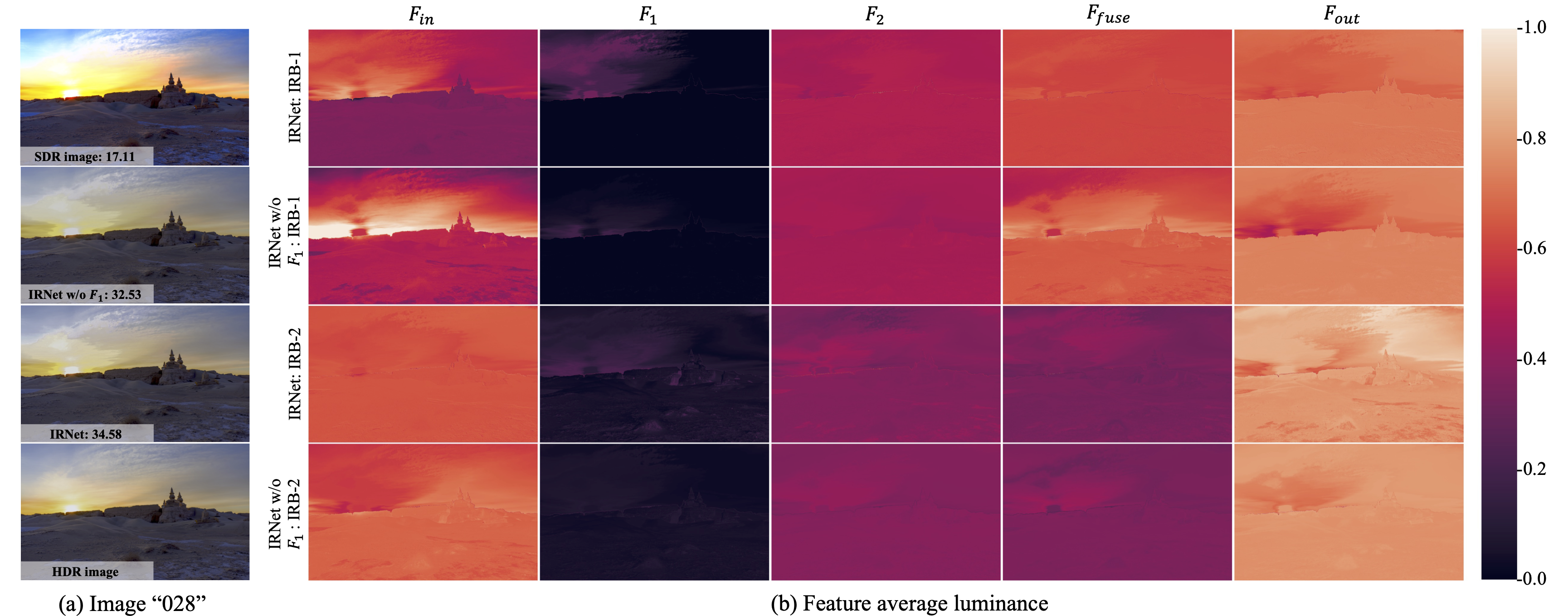}
\vspace{-2mm}
\caption{
(a) \textbf{Visual comparison} of the SDR image, the ITM results by our IRNet and our IRNet w/o $F_1$, and the corresponding HDR image. The PSNR (dB) results are provided for reference.
(b) \textbf{Visualization of the mean feature maps} output by the $1$-st and $2$-nd IRB blocks of our IRNet and our IRNet w/o using $F_1$ on ITM. The mean feature maps are averaged along the channel dimension and also normalized for better visualization.
Denote $M^{1}$ and $M^{2}$ as the two corresponding mean feature maps of IRNet and IRNet w/o $F_1$, respectively, while $M^{1}_{max}$ (or $M^{2}_{max}$) and $M^{1}_{min}$ (or $M^{2}_{min}$) as the maximum and minimum values of $M^{1}$ (or $M^{2}$), respectively. Denote $\min(\cdot,\cdot)$ and $\max(\cdot,\cdot)$ as the minimum and maximum values of the two values in the bracket, respectively. We normalize the mean feature map by $M^{i}=(M^{i}-\min(M^{1}_{min},M^{2}_{min}))/\max(M^{1}_{max}-M^{1}_{min},M^{2}_{max}-M^{2}_{min})$.
}
\label{fig:feature}
\centering
\vspace{-3mm}
\end{figure*}

For promising ITM performance, the \sota ITM methods~\cite{endo,eilertsen2017hdr,Marnerides2018Expandnet,santos2020single,9447972} suffer from complex network backbones with huge parameter amounts and computational costs. To implement efficient ITM, in this paper, we propose to develop a simple, lightweight, and efficient Improved Residual Network (IRNet) by slightly modifying the residual block~\cite{he2016deep}. As shown in Figure~\ref{fig:arch} (b), in the proposed Improved Residual Block (IRB), we concatenate the intermediate feature map $F_1$ extracted after the LeakyReLU with the fused feature of $F_{in}$ and $F_{2}$ (more details will be presented in \S\ref{sec:irnet}). Our IRNet shows clear improvements, especially in the bright area near the sun, over that without using the feature map $F_1$ on the image ``028'' from the HDRTV1K test set~\cite{chen2021new}, as shown in Figure \ref{fig:luminance&1-D} (b). Along the highlighted lines,
the \textcolor{green}{green} line of our IRNet enjoys closer approximation to the \textcolor{blue}{blue} line of the ``Ground Truth'' HDR image than the \textcolor{red}{red} line of our IRNet without using the intermediate feature $F_1$ (denoted as ``IRNet w/o $F_1$''). In Figure \ref{fig:luminance&1-D} (c), we plot the ratios of luminance values of the highlighted lines by our IRNet and ``IRNet w/o $F_1$'', which also validates that our IRNet achieves better approximation to the ``Ground Truth'' than the IRNet without $F_1$. This validates the effectiveness of our IRB over the residual block for ITM.

Adaptive luminance extension is also important for the ITM task. For this goal, many joint SR-ITM methods~\cite{Kim_Oh_Kim_2020,9008274,xu2023fdan} performed image or feature decomposition to extract and fuse multi-scale feature maps. However, these ITM networks with decomposition techniques often suffer from complex network structures with heavy computational costs (Table~\ref{tab:ITM_sotacomparison}). For efficiency consideration, we design our IRNet as a simple and lightweight network by employing the popular residual block~\cite{he2016deep} as a proper backbone for our IRNet. The promising results in Figure~\ref{fig:luminance&1-D} (b) by our IRNet without $F_1$ motivates us to boost it for better ITM performance.

\subsection{Proposed Improved Residual Network}

\label{sec:irnet}
The motivation of designing our IRNet is to fully exploit the effectiveness of multi-scale feature fusion at both intra-block (within our IRB) and inter-block (between different IRBs) levels. To be specific, our IRNet first extracts the initial feature map using a $1\times1$ convolution layer (instead of $3\times3$ one to reduce the parameter amounts).  Then, empowered by the intra-block multi-scale feature fusion, $n$ cascaded Improved Residual Blocks (IRBs) are responsible for fine-grained feature extraction. Details of our IRB will be introduced later. To boost the utilization of the refined features, each IRB is followed by a Contrast-aware Channel Attention (CCA) layer~\cite{CCA}. We also explore the multi-scale feature fusion at inter-block level by adding a residual connection between each paired IRB block and the CCA layer~\cite{CCA}, which further facilitates the information flow. We now elaborate the components of our IRNet.

\noindent
\textbf{Improved Residual Block (IRB)}. The proposed IRB block is built upon the residual block~\cite{he2016deep}, which achieves great success in many computer vision tasks~\cite{chen2021new,9156371}. As shown in Figure~\ref{fig:arch} (a), the residual block~\cite{he2016deep} contains two $3 \times 3$ convolution layers with an activation function (here we replace ReLU by LeakyReLU) between them, the output feature is added to the input feature $F_{in}$ and activated by LeakyReLU.

Built upon the residual block, our IRB block is designed to keep our IRNet as simple as possible with better ITM performance. This is feasible by fully exploiting the multi-layer feature maps within the IRB block. To this end, given the input feature $F_{in}\in\mathbb{R}^{H\times W\times C}$, our IRB first refines it by a $3\times3$ convolution layer and a LeakyReLU activation function. The extracted feature $F_1\in\mathbb{R}^{H\times W\times C/2}$ is further refined in our IRB by a second $3\times3$ convolution layer to output the feature $F_{2}\in\mathbb{R}^{H\times W\times C}$:
\begin{gather}
    F_1 = \operatorname{LeakyReLU} \left(\operatorname{Conv_{3 \times 3}}\left(F_{in}\right)\right),\\
    F_2 = \operatorname{Conv_{3\times 3}}\left(F_1 \right).
\end{gather}
Then our IRB uses a skip connection and a $\operatorname{Conv_{1\times1}}$ to fuse $F_{in}$ and $F_{2}$ and obtain the fusion feature $F_{fuse}$:
\begin{gather}
    F_{fuse} = \operatorname{Conv_{1\times1}}\left(F_{in}+F_{2}\right).
\end{gather}
Finally,  our IRB explicitly concatenates the intermediate feature $F_1$ 
with the fusion feature $F_{fuse}$
to produce the output feature $F_{out}$ as follows:
\begin{gather}
F_{out}=\operatorname{Conv_{1\times1}}\left(\operatorname{Concat}\left(F_{fuse},F_{1}\right)\right).
\end{gather}
We visualize the structure of our IRB block in Figure \ref{fig:arch} (b).

Compared with the original residual block, our IRB well extracts and utilizes the multi-layer features, which correspond to spatially adaptive luminance areas for ITM. As shown in Figure~\ref{fig:feature} (a), compared with the IRNet w/o $F_1$, our IRNet restores the luminance of HDR image closer to the ground truth, especially in the highlight regions. Even though popular encoder-decoder frameworks like U-net~\cite{unet} or Uformer~\cite{uformer} can be utilized here to extract strong multi-scale features, this would bring significant growth on parameter amounts and computational costs~\cite{chen2021hdrunet}. Through a simple modification to the residual block, the proposed IRB serves as a lightweight building block in our IRNet for efficient ITM performance.

The mean feature map along the channel dimension could reflects the luminance information of that feature~\cite{xu2023fdan}. In Figure~\ref{fig:feature} (b), we visualize the mean feature maps of $F_{in}$, $F_1$, $F_2$, $F_{fuse}$, and $F_{out}$ extracted by our IRNet and ``IRNet w/o $F_1$''. One can see that the mean feature map of $F_1$ extracted by our IRNet exhibits higher luminance in the sky area around the sun than that of ``IRNet w/o $F_1$''. Due to the lack of luminance information by the intermediate feature $F_1$, ``IRNet w/o $F_1$'' produces stronger contrasts at the input feature $F_{in}$ of IRB blocks and darker luminance around the sun in the output feature $F_{out}$, than our IRNet using $F_1$ in our IRB block.

\noindent
\textbf{Contrast-aware Channel Attention (CCA)}. To preserve image details, we utilize a CCA layer~\cite{CCA} after each IRB block. As shown in Figure~\ref{fig:arch} (c), the CCA layer~\cite{CCA} consists of contrast computation, two $1\times1$ convolution layers interleaved with a ReLU function, a sigmoid function, and a skip connection between the input and output features to help gradient propagation. Given the input $X=[x_1,...,x_C]\in\mathbb{R}^{H\times W\times C}$, the contrast is computed as follows:
\begin{align}
    z_c = H_{GC}(x_c)
    & = \sqrt{\frac{1}{HW}\sum_{(i,j)\in x_c} (x_c^{i,j}-\sum_{(i,j)\in x_c} x_c^{i,j})^{2}} + \notag \\
    & \frac{1}{HW}\sum_{(i,j)\in x_c} x_c^{i,j}, c=1,....,C.
\end{align}
After the $i$-th ($i=1,...,n-1$) IRB block and CCA layer~\cite{CCA}, the output feature is added to the input feature $F_{in}^{i}$ by a skip connection, and $F_{in}^{n+1}$ is the final feature that will be inputted to the next convolution layers as follows:
\begin{gather}
\label{step}
F_{in}^{i+1}=F_{in}^{i}+\operatorname{CCA}\left(\operatorname{IRB}\left(F_{in}^{i}\right)\right).
\end{gather}

After extracting $n$ scales of fine-grained feature maps, we concatenate them for multi-scale feature fusion, which is implemented by a sequence of $1\times1$ convolution layer, a LeakyReLU activation function, and a $3\times3$ convolution layer. Finally, we reconstruct the output HDR image using a $3\times3$ convolution layer. The overall architecture of the proposed IRNet is shown in Figure \ref{fig:arch} (d).

To apply the proposed IRNet to the joint SR-ITM task, we further add a Pixel Shuffle operation~\cite{shi2016real} after the final $3\times3$ convolution layer of our IRNet to make it feasible for super-resolution. The Pixel-Shuffle contains two $3\times3$ convolution layers interleaved with a ReLU function. The first convolution layer reduces the channel dimension of the feature map from $C$ to $3s^2$, where $s$ is the upsampling factor, while the second convolution layer reconstructs the $3$-channel HR-HDR image via upsampling the feature map by a factor of $s$.

\subsection{Implementation Details} 
Here, we set the channel dimension of the feature map $F_{in}$ as $C=64$. The number of IRB blocks $n$ is set as $n=2$ for the ITM task and $n=5$ for the joint SR-ITM task. We use Kaiming initialization~\cite{he2016deep} to initialize the parameters of our IRNet.
To optimize these parameters, we adopt Adam optimizer~\cite{kingma2014adam} with $\beta_1 = 0.9$ and $\beta_2 = 0.999$ to minimize an $\ell_{1}$ loss function. The learning rate $\eta$ is initialized as $5\times10^{-4}$ and degrades to $1\times10^{-11}$ by cosine annealing schedule with warm restart~\cite{loshchilov2016sgdr} in every 60 epochs. The batch size is set as 16. We train the models of our IRNet for 200 epochs on an NVIDIA V100 GPU with 32GB memory.

\section{Experiments}
\label{sec:experiments}

In this section, we evaluate the performance of comparison methods on the ITM and joint SR-ITM tasks. We first introduce the used datasets and metrics. Then we present the the comparison results on ITM and joint SR-ITM, respectively. Finally, we conduct a series of ablation experiments to study the components of our IRNet.


\begin{figure*}[t]
\centering
\includegraphics[width=13cm]{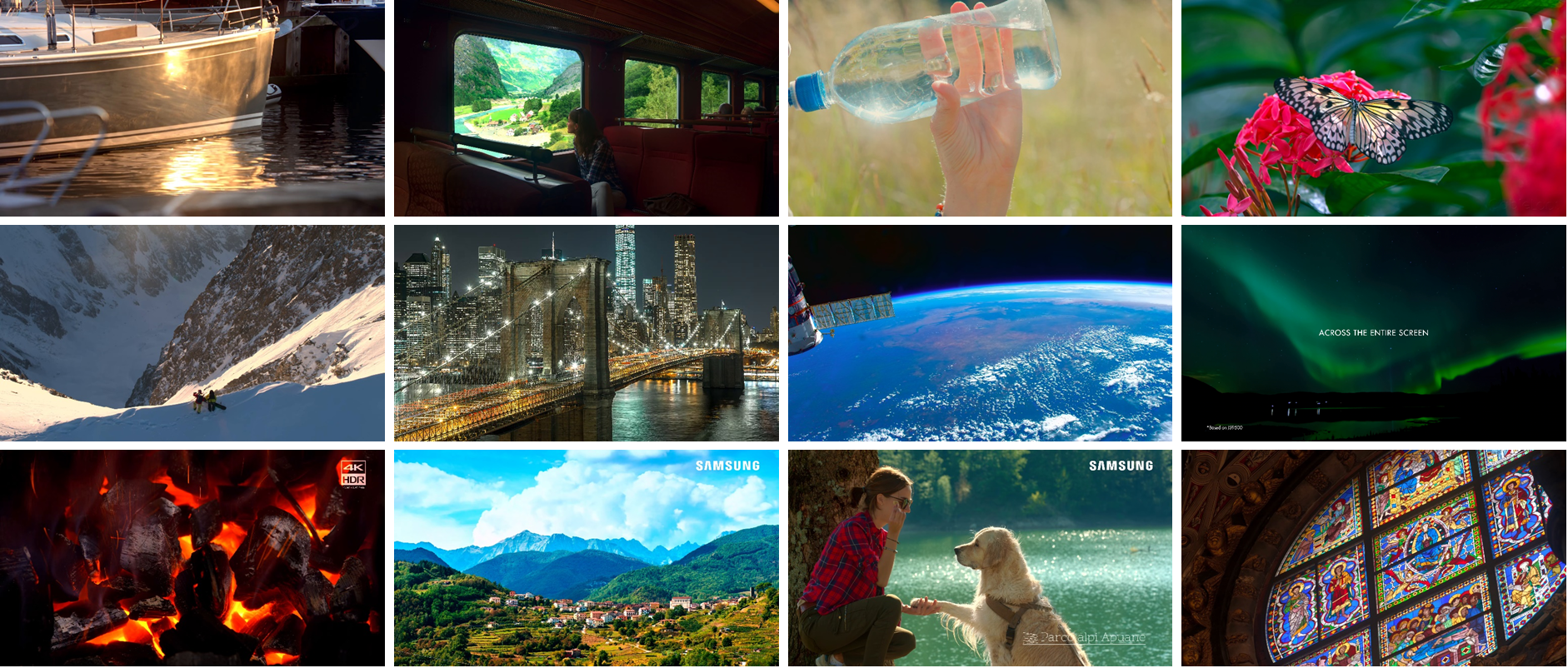}
\vspace{-3mm}
\caption{\textbf{12 typical scenes of our ITM-4K dataset} for method evaluation on the ITM task.}
\label{fig:new}
\end{figure*}

\subsection{Dataset and Metrics}
\label{dataset}

\noindent
\textbf{Training set}. In our experiments, we use the recently published HDRTV1K dataset~\cite{chen2021new} to evaluate the comparison methods. This dataset contains 1,235 pairs of 8-bit SDR and 10-bit HDR images for training and 117 pairs of images for testing. We crop each image in the training set into 30 $ 256\times256$ image patches. For data augmentation, we randomly flip the cropped patches horizontally or vertically, rotate these patches by 90°, 180°, or 270°.
To perform joint SR-ITM on the HDRTV1K dataset, which is originally developed only for ITM, we downsample the SDR images by a factor of $s=4$ to obtain the low-resolution (LR) SDR images, similar to~\cite{9008274}. The high-resolution (HR) and HDR images from the HDRTV1K dataset can still be used as the training targets.

\begin{figure}[!t]
\centering
\vspace{-0.3cm} 
\includegraphics[width=0.6\textwidth]{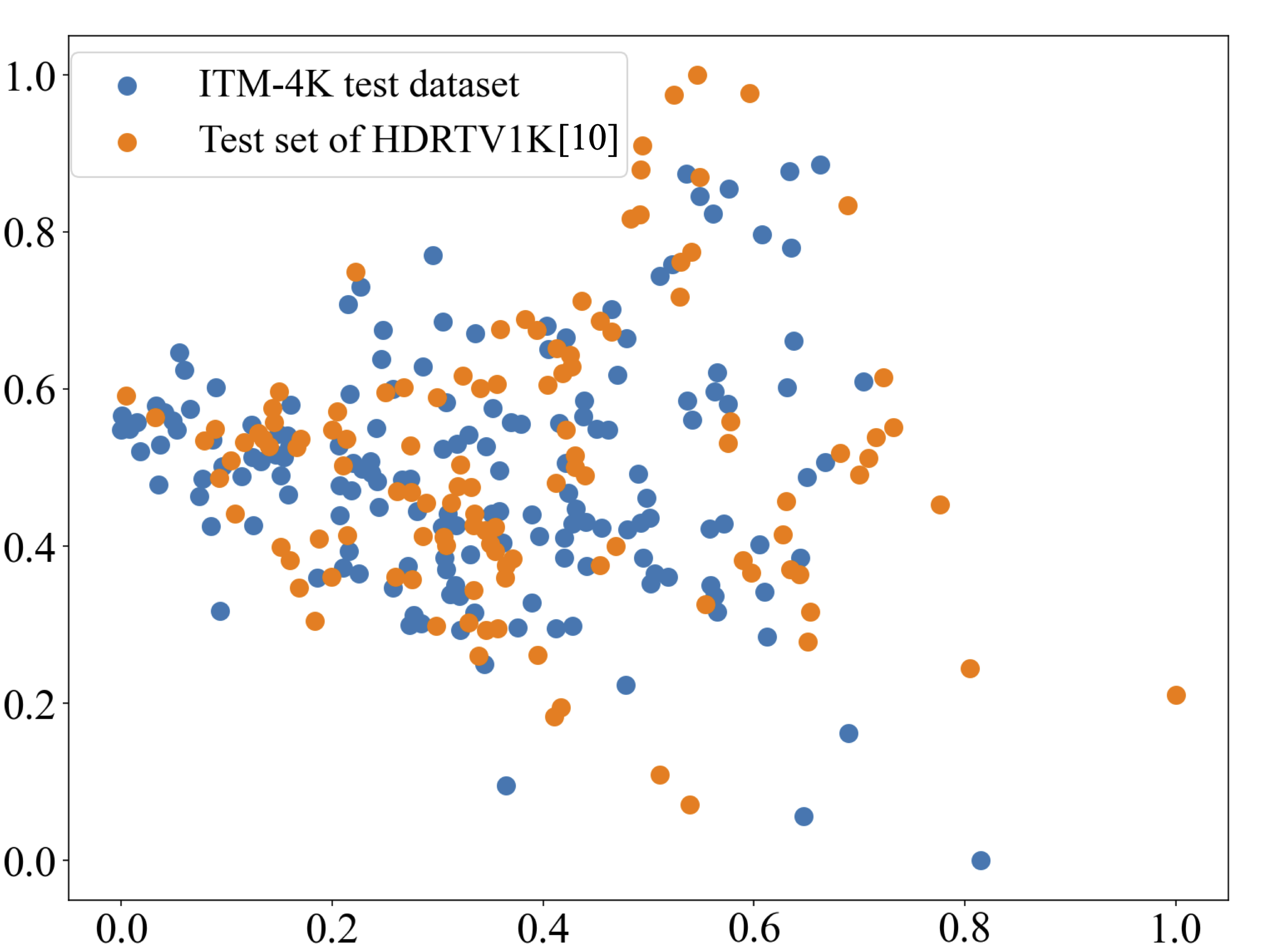}
\caption{ {\bfseries Data distribution of our ITM-4K test set and the test set of HDRTV1K \cite{chen2021new}.} t-SNE \cite{Maaten2008VisualizingDU} is adopted for dimension reduction on images, and the final values are normalized to $[0,1]$.}\label{fig:new_compare}
\vspace{-4mm}
\end{figure}

\noindent
\textbf{Test sets}. On the ITM task, we evaluate the comparison methods on three datasets: the test set of HDRTV1K \cite{chen2021new}, our newly collected ITM-4K dataset (for high-resolution images), and the test set in \cite{9008274}. On the joint SR-ITM task, we evaluate the comparison methods on the test set of HDRTV1K~\cite{chen2021new}. The details of these test sets are summarized as follows:
\begin{itemize}
\item \textbf{HDRTV1K}~\cite{chen2021new} contains 117 test SDR images of size $3840\times2160\times3$, with paired HDR images. For joint SR-ITM, we downsample the SDR images by a factor of 4 to generate the LR-SDR test images.
\item \textbf{ITM-4K} contains 160 pairs of SDR and HDR images of size $3840\times2160\times3$. These images are extracted from 9 HDR10 videos collected from \href{https://4kmedia.org}{4kmedia.org} and do not overlap with HDRTV1K \cite{chen2021new}. The corresponding SDR videos are generated through YouTube similar to~\cite{9008274}. We display 12 typical scenes from the 160 test images in Figure~\ref{fig:new}. In Figure~\ref{fig:new_compare}, we also visualize the distribution of the 160 SDR images in our ITM-4K dataset and the 117 SDR test images in HDRTV1K~\cite{chen2021new} using $t$-SNE~\cite{Maaten2008VisualizingDU}. One can see that our ITM-4K dataset contains diverse scenes similar yet supplementary to the test set of HDRTV1K~\cite{chen2021new}.
\item \textbf{The test set in~\cite{9008274}}. This dataset contains 28 test images, 12 of which are overlapped with the training set of \textbf{HDRTV1K}~\cite{chen2021new} and the test set of our \textbf{ITM-4K}. Thus, we use the remaining 16 images to evaluate the ITM methods. Note that although this dataset is used for joint SR-ITM task, the test set provides the SDR images of the same sizes with the corresponding HDR images, which can be used to evaluate ITM methods. We do not use this test set for the joint SR-ITM task due to its overlap with the training set of HDRTV1K~\cite{chen2021new}.

\end{itemize}

\noindent
\textbf{Metrics}. We evaluate the performance of different methods on ITM and joint SR-ITM in terms of PSNR, SSIM~\cite{wang2004image}, LPIPS~\cite{zhang2018unreasonable}, and HDR-VDP3~\cite{mantiuk2011hdr}. PSNR is used to evaluate the closeness of the output image to the corresponding ground truth image. SSIM~\cite{wang2004image} and LPIPS~\cite{zhang2018unreasonable} evaluate the structural and perceptual similarity, respectively, of the output image to the corresponding ground truth image. HDR-VDP3~\cite{mantiuk2011hdr} is a widely used metric to evaluate the quality of HDR images~\cite{chen2021new,Chen2022HDRre}, and we use its prediction of ``quality'' (Q) here.

\subsection{Results on Inverse Tone Mapping}
\label{ITM-result}

\noindent
\textbf{Comparison methods}. 
For our IRNet, we set $n=2$ and $C=64$, and denote it as ``IRNet-2 (64c)''. 
We compare it with four ITM methods of HDRNet \cite{gharbi2017deep}, CSRNet \cite{he2020conditional}, Ada-3DLUT \cite{zeng2020learning}, and HDRTVNet \cite{chen2021new}. The methods of Pixel2Pixel~\cite{isola2017image} and CycleGAN~\cite{zhu2017unpaired} are also evaluated as two generative baselines for ITM. As suggested in~\cite{chen2021new}, we also modify the joint SR-ITM methods of Deep SR-ITM~\cite{9008274} and JSI-GAN~\cite{Kim_Oh_Kim_2020} for the ITM task, by setting the stride of the first convolution layer as 2 to make them feasible for the ITM task. This manner reduces their computational costs while not degrading the ITM performance.

\noindent
\textbf{Objective results}. The comparison results on the test set of HDRTV1K~\cite{chen2021new} are summarized in Table \ref{tab:ITM_sotacomparison}. One can see that our IRNet-2 (64c) outperforms the second best method, \ie, AGCM+LE, by 0.59dB, 0.0011, and 0.3 in terms of PSNR, SSIM, and LPIPS, respectively. Note that our IRNet-2 (64c) has 134.73K parameters, fewer than all the other comparison methods except CSRNet (36.49K) and AGCM (35.25K). But these two methods suffer from clear performance gap to our IRNet-2 (64c) in terms of all evaluation metrics. On HDR-VDP3, our method is slightly (0.03) lower than the best method AGCM+LE. But AGCM+LE requires 1410K parameters, 6228.31G FLOPs, and 3114.09G MACs to process a 4K-resolution SDR image at a speed of 691.30ms, much larger than those of our IRNet-2 (64c).
Besides, our IRNet-1 (48c), \ie, the IRNet with a single IRB block and $C=48$, only needs 49.3K parameters to achieve competitive results with the second best method of AGCM+LE.

We further evaluate our IRNet-2 and other methods on our ITM-4K dataset and the 16 SDR images in the test set of~\cite{9008274}. As shown in Table \ref{tab:ITM_new_table}, our IRNet-2 (64c) still achieves better results than  other comparison methods on PSNR and HDR-VDP3. In summary, our IRNet achieves efficient ITM performance with a lightweight backbone.

\begin{table*}[t]\footnotesize
    \caption{\textbf{Comparison of parameter amounts, FLOPs on 4K-resolution images, running time on 4K images, PSNR, SSIM, LPIPS and HDR-VDP3 (quality) for ITM tasks on the test set of HDRTV1K~\cite{chen2021new}}. The {\color{red} \textbf{Red}}, {\color{blue} \textbf{blue}} and \textbf{bold} texts indicate the best, the second best, and the third best results, respectively.}
    \label{tab:ITM_sotacomparison}
    \centering
    \renewcommand{\arraystretch}{1.2}
    \setlength{\tabcolsep}{1.6mm}

    \begin{NiceTabular}{r||r|r|r|r||r|r|r|r}
        \Xhline{1.4pt}
            \multirow{2}{*}{Method} & Params.$\downarrow$ & FLOPs$\downarrow$ & MACs$\downarrow$ & Time$\downarrow$ & PSNR$\uparrow$ & \multirow{2}{*}{SSIM$\uparrow$} & LPIPS$\downarrow$ & HDR- \\
            & (K) & (G) & (G) & (ms) & (dB) & & (\%) & VDP3$\uparrow$ \\
            \hhline{=||=|=|=|=||=|=|=|=}
            \rowcolor{gray!10}
            Pixel2Pixel \cite{isola2017image} & 11380.00 & 5227.49 & 2613.07 & 188.20 & 28.33 & 0.9177 & 8.94 & 6.91 \\ 
            CycleGAN \cite{zhu2017unpaired} & 11380.00 & 14380.38 & 7188.66 & 1633.38 & 20.69 & 0.7842 & 17.41 & 6.41 \\
        \hhline{-||----||----}
            \rowcolor{gray!10}
            HDRNet \cite{gharbi2017deep} & 482.00 & 1.84 & 0.92 & 44.50 & 35.42 & 0.9670 & 2.94 & 8.52\\
            CSRNet \cite{he2020conditional} & 36.49 & 106.28 & 52.84 & 78.14 & 35.21 & 0.9646 & 3.13 & 8.44 \\
            \rowcolor{gray!10}
            Ada-3DLUT \cite{zeng2020learning} & 594.01 & 1.57 & 0.82 & 4.55 & 36.22 & 0.9662 & 3.20 & 8.43\\
        \hhline{-||----||----} 
            Deep SR-ITM \cite{9008274} & 2500.00 & 10449.60 & 5224.80 & 777.95 & 30.79 & 0.8992 & 10.38 & 6.89 \\
            \rowcolor{gray!10}
            JSI-GAN \cite{Kim_Oh_Kim_2020} & 1454.24 & 6133.98 & 3066.99 & 1780.66
 & 36.32 &  0.9610 & 3.84 & 8.01\\
        \hhline{-||----||----}
            AGCM \cite{chen2021new} & 35.25 & 77.05 & 38.53 & 99.24 & 36.88 & 0.9655 & 3.21 & 8.46 \\
            \rowcolor{gray!10}
            AGCM+LE \cite{chen2021new} & 1410.00 & 6228.31 & 3114.09 & 691.30 & \textbf{37.61} & \textbf{0.9726} & \textbf{3.06} & {\color{red} \textbf{8.61}} \\
            HDRTVNet \cite{chen2021new} & 37200.00 & 14102.81 & 7051.34 & 1513.43 & 37.21 & 0.9699 & 5.76 & \textbf{8.57} \\
        \hhline{-||----||----}
          \rowcolor{gray!10}
          \textbf{IRNet-1 (48c)} & 49.30 & 810.20 & 404.10 & 166.91 & {\color{blue} \textbf{37.82}} & {\color{blue} \textbf{0.9730}} & {\color{red} \textbf{2.62}} &  8.54\\          
          \textbf{IRNet-2 (64c)} & 134.73 & 2211.49 & 1104.15 & 398.33 & {\color{red} \textbf{38.20}} & {\color{red} \textbf{0.9737}} & {\color{blue} \textbf{2.68}} &  {\color{blue} \textbf{8.58}} \\
        \hline
    \end{NiceTabular}
    \vspace{-3mm}
\end{table*}

\begin{table*}\footnotesize

    \caption{\textbf{Comparison of parameter amounts, PSNR, SSIM, LPIPS and HDR-VDP3 (quality) for ITM tasks on our ITM-4K test dataset and the test set of \cite{9008274}}. {\color{red} \textbf{Red}}, {\color{blue} \textbf{blue}} and \textbf{bold} texts indicate the best, second best, and third best results, respectively.
    }
    \label{tab:ITM_new_table}
    \centering
    \renewcommand{\arraystretch}{1.2}
    \setlength{\tabcolsep}{2.6mm}

    \begin{NiceTabular}{c||r||c|c|c|c}
        \Xhline{1.4pt}
            Test dataset & Method & PSNR (dB)$\uparrow$ & SSIM$\uparrow$ & LPIPS (\%)$\downarrow$ & HDR-VDP3$\uparrow$\\
        \hhline{=||=||=|=|=|=}
            \rowcolor{gray!10}
            \cellcolor{white} & Pixel2Pixel \cite{isola2017image} & 28.27 & 0.9090 & 9.01 & 6.99 \\ 
            & CycleGAN \cite{zhu2017unpaired} & 21.47 & 0.7724 & 16.25 & 6.38 \\
            \hhline{~||-||----}
            \rowcolor{gray!10}
            \cellcolor{white} & HDRNet \cite{gharbi2017deep} & 33.86 &  \textbf{0.9420}  & \textbf{4.94} & 8.13  \\
            & CSRNet \cite{he2020conditional} & 33.65 & 0.9379 & {\color{blue} \textbf{4.93}} & 8.13 \\
            \rowcolor{gray!10}
            \cellcolor{white} & Ada-3DLUT \cite{zeng2020learning} & \textbf{34.38} & 0.9399 & {\color{red} \textbf{4.68}} & \textbf{8.17} \\
            \hhline{~||-||----}
            & Deep SR-ITM \cite{9008274} & 30.27 & 0.8805 & 9.18 & 6.85 \\
            \rowcolor{gray!10}
            \cellcolor{white} & JSI-GAN \cite{Kim_Oh_Kim_2020} & 33.57 & 0.9365 & 6.27 & 7.24 \\
            & AGCM+LE \cite{chen2021new} & {\color{blue} \textbf{34.73}} & {\color{blue} \textbf{0.9453}} & 5.94 & {\color{blue} \textbf{8.21}}\\
            \hhline{~||-||----}
            \rowcolor{gray!10}
            \cellcolor{white} \multirow{-9}{*}{ITM-4K} & \textbf{IRNet-2 (64c)} & {\color{red} \textbf{35.24}} & {\color{red} \textbf{0.9459}} & 5.48 & {\color{red} \textbf{8.23}}  \\
        \hhline{=||=||=|=|=|=}
            \rowcolor{gray!10}
            \cellcolor{white} & Pixel2Pixel \cite{isola2017image} & 26.05 & 0.9200 & 5.86 & 7.01 \\
            & CycleGAN \cite{zhu2017unpaired} & 21.54 & 0.7255 & 20.77 & 5.85 \\
        \hhline{~||-||----}
            \rowcolor{gray!10}
            \cellcolor{white} & HDRNet \cite{gharbi2017deep} & 33.03 & \textbf{0.9579} & {\color{blue} \textbf{2.73}} & {\color{blue} \textbf{8.21}} \\
            & CSRNet \cite{he2020conditional}  & 32.09 & 0.9559 & \textbf{2.83} & 8.15 \\
            
            \rowcolor{gray!10}
            \cellcolor{white} & Ada-3DLUT \cite{zeng2020learning} & \textbf{33.08} & 0.9365 & {\color{red} \textbf{2.51}} & 8.12 \\ 
        \hhline{~||-||----}
            & Deep SR-ITM \cite{9008274} & 27.49 & 0.8752 & 9.32 & 5.82\\
            \rowcolor{gray!10}
            \cellcolor{white} & JSI-GAN \cite{Kim_Oh_Kim_2020} & 31.36 & 0.9539 & 4.21 & 7.16 \\
            & AGCM+LE \cite{chen2021new} & {\color{blue} \textbf{33.49}} & {\color{red} \textbf{0.9614}} & 3.13 & {\color{red} \textbf{8.22}} \\
        \hhline{~||-||----}
            \rowcolor{gray!10}
            \cellcolor{white} \multirow{-9}{*}{test set in \cite{9008274}} & \textbf{IRNet-2 (64c)} & {\color{red} \textbf{33.85}} & {\color{blue} \textbf{0.9609}} & 2.90 & \textbf{8.20}  \\
        \hline
    \end{NiceTabular}
    \vspace{-3mm}
\end{table*}

\begin{figure*}[t]
    \centering
    \vspace{-2mm}
\includegraphics[width=13cm]{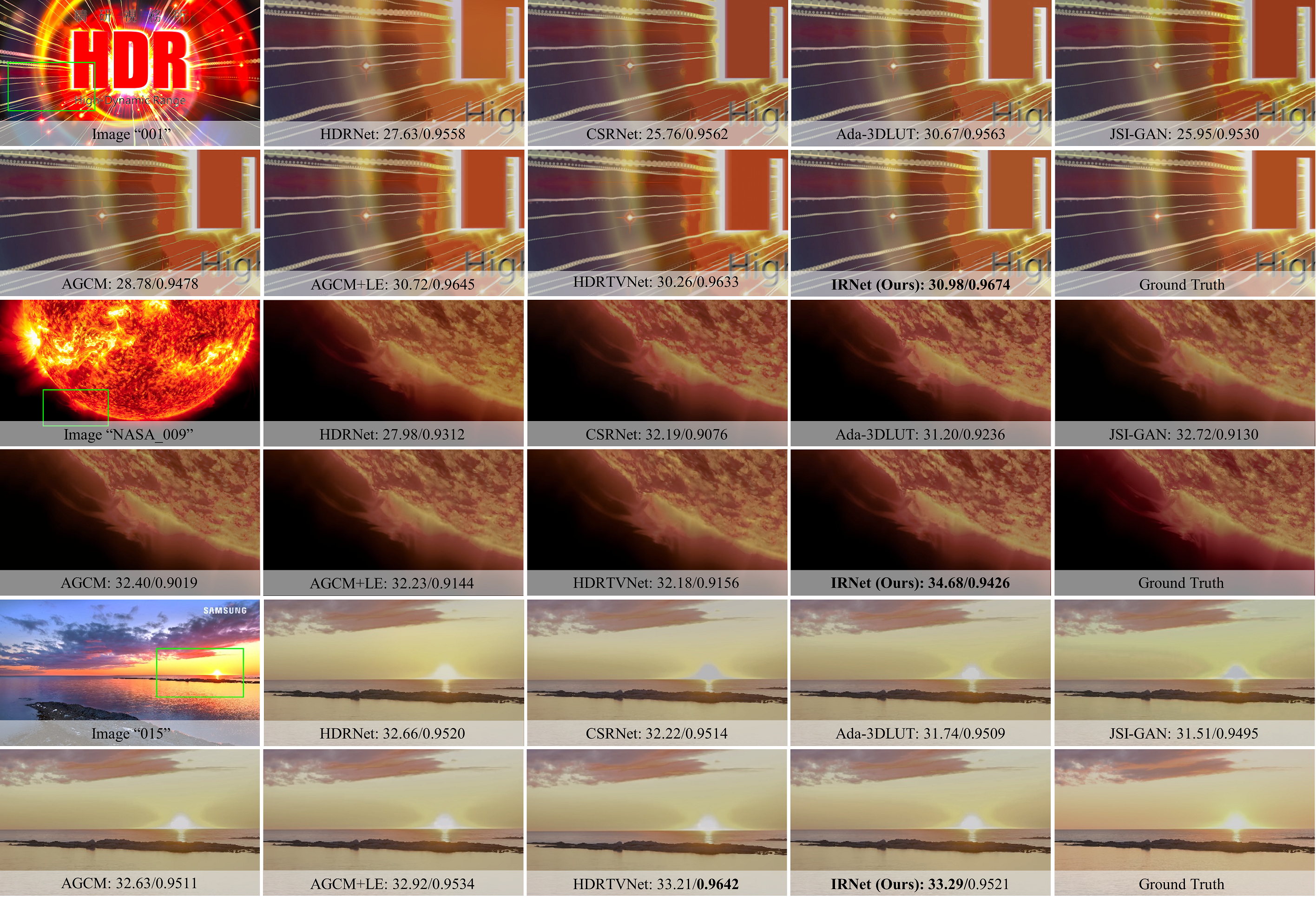}
\vspace{-3mm}
    \caption{ {\bfseries Visual quality, PSNR (dB) and SSIM results by different ITM methods on the HDRTV1K test set~\cite{chen2021new} (1-st row), the test set of \cite{9008274} (2-nd row), and our ITM-4K test dataset (3-rd row)}. For each image, we enlarge a cropped region for better visualization.
    The parameter amounts, computational costs, and speed are summarized in Table 
    \ref{tab:ITM_sotacomparison}. 
    } \label{fig:com}
    \vspace{-2.5mm}
\end{figure*}

\noindent
\textbf{Visual quality} is an important criterion to evaluate the performance of ITM methods, since human are the final reviewers of the image quality. For the purpose of visualization, the HDR images are generated from HDR10 videos and stored in the 16-bit PNG format. The comparison results of visual quality by different methods on three test sets are shown in Figure~\ref{fig:com}. We observe that most comparison methods suffer from a certain degree of color bias, especially near the light source. Our IRNet achieves closer results to the ground truth images than other methods, with more correct colors and color contrasts. In addition, our IRNet achieves better PSNR and SSIM results than the other comparison methods. All these results demonstrate that our IRNet is very effective on ITM.

\noindent
\textbf{Running speed} is the actual wall-clock time of evaluating model efficiency on SDR images. We calculate the running time of comparison methods on 4K-resolution ($3840\times2160\times3$) images. As shown in Table \ref{tab:ITM_sotacomparison}, our IRNet-2 (64c) is faster than the second and third best methods, \ie, AGCM+LE and HDRTVNet, by a gap of 292.97ms and 1115.10ms, respectively. Meanwhile, IRNet-1 (48c) reduces the running time of IRNet-2 (64c) from 398.33ms to 166.91ms with guaranteed performance. Although faster than our IRNet-2, the methods of HDRNet, CSRNet, Ada-3DLUT, and AGCM suffer from obvious performance degradation on quantitative metrics.



\begin{figure*}[ht]
    \centering \includegraphics[width=13cm]{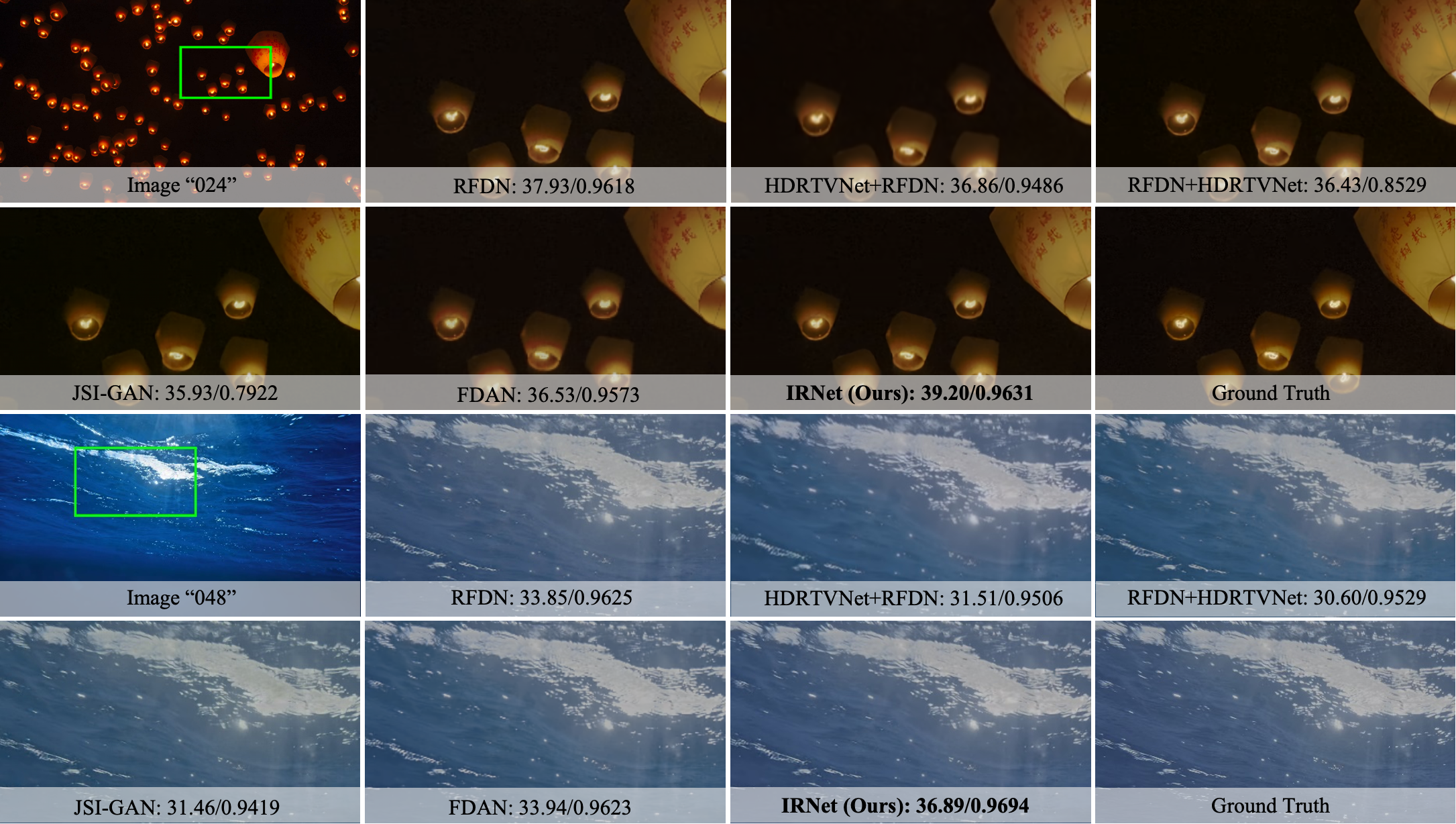}
    \vspace{-2mm}
    \caption{{\bfseries Visual quality, PSNR (dB) and SSIM results by different joint SR-ITM methods on two SDR images from \cite{chen2021new}}. For each image, we enlarge a cropped region for better visualization. The parameter amounts, computational costs, and speed are summarized in Table \ref{tab:SR-ITMtable}. 
    }\label{com_SRITM}
\vspace{-3mm}
\end{figure*}

\subsection{Results on Joint SR-ITM}
\label{SRITM-result}

\noindent
\textbf{Comparison methods}. Here, we set $n=5$ and $C=64$ in our IRNet, and denote it as ``IRNet-5 (64c)''. We compare it with two \sota SR methods, \ie, EDSR~\cite{edsr} and RFDN~\cite{Liu2020}, two cascaded two stage SR-ITM methods, \ie, ``HDRTVNet+RFDN'' (sequentially performing ITM by HDRTVNet and SR by RFDN) and ``RFDN+HDRTVNet'' (vice versa), and three joint SR-ITM methods, \ie, Deep SR-ITM \cite{9008274}, JSI-GAN \cite{Kim_Oh_Kim_2020}, and FDAN~\cite{xu2023fdan}. For the cascaded SR-ITM methods, we choose RFDN~\cite{Liu2020} and HDRTVNet\cite{chen2021new} since they are \sota methods on SR and ITM, respectively.

\noindent
\textbf{Objective results}. The comparison of numerical results are summarized in Table \ref{tab:SR-ITMtable}. It can be seen that
the two SR methods still achieve reasonable performance in terms of objective metrics. By first performing SR and then ITM, the cascaded method achieves better results on image quality metrics, but requires heavy computational costs, \eg, 14783.55G FLOPs and 7391.58G MACs to process an LR-SDR image of size $960\times540$. Of course, first performing ITM and then SR significantly reduces the computational costs, and the performance on evaluation metrics suffers from a huge degradation as well. Besides, compared with joint SR-ITM methods, \ie, Deep SR-ITM, JSI-GAN and FDAN, our IRNet-5 (64c) achieves the best PSNR results (0.38dB higher than the second best method ``RFDN+HDRTVNet'') and comparable results on other metrics, but with small requirements on parameter amounts, computational costs, and inference time. These observations demonstrate that our IRNet is a lightweight and efficient backbone that can achieve \sota performance on the joint SR-ITM task.

\begin{table*}[pt]\footnotesize
\caption{\textbf{Comparison results of evaluation metrics by different methods on the joint SR-ITM task with a scale of 4 on the test set of HDRTV1K \cite{chen2021new}}. The FLOPs, MACs, and time (ms) are test on images of size $ 960 \times 540 \times 3 $. On the metrics of PSNR, SSIM, LPIPS and HDR-VDP3 (quality), {\color{red} \textbf{Red}} and {\color{blue} \textbf{blue}} texts indicate the best and second best results, respectively.
}
\renewcommand{\arraystretch}{1.2}
    \setlength{\tabcolsep}{0.95mm}
    \centering
    \begin{NiceTabular}{c||r|r|r|r||r|r|r|r}
        \Xhline{1.4pt}
             \multirow{2}{*}{Method} & Params.$\downarrow$ & FLOPs$\downarrow$ & MACs$\downarrow$ & Time$\downarrow$ & PSNR$\uparrow$ & SSIM$\uparrow$ & LPIPS$\downarrow$ & HDR- \\
            & (K) & (G) & (G) & (ms) & (dB) &  & (\%) & VDP3$\uparrow$ \\
        \hhline{=||=|=|=|=||=|=|=|=}
                \rowcolor{gray!10}
                EDSR \cite{edsr}  & 43090.00 & 52102.13 & 26051.00 & 4737.07 & 18.87 & 0.6866 & 29.09 & 5.54 \\
                
                RFDN \cite{Liu2020} & 660.16 & 680.74 & 340.24 & 79.37 & 33.07 & 0.9111 & 11.89 & 7.48 \\
            \hhline{-||----||----}
                \rowcolor{gray!10}
                HDRTVNet\cite{chen2021new}+RFDN\cite{Liu2020} & 37860.00 & 881.42 & 440.71 & 154.95 & 32.29 & 0.8982 & 19.12 & 7.24 \\
               
                RFDN\cite{Liu2020}+HDRTVNet\cite{chen2021new} & 37860.00 & 14783.55 & 7391.58 & 1592.80 & {\color{blue} \textbf{33.36}} & {\color{red} \textbf{0.9178}} & {\color{red} \textbf{10.83}} & {\color{red} \textbf{7.61}}\\
            \hhline{-||----||----}
                \rowcolor{gray!10}
                Deep SR-ITM \cite{9008274} & 2950.00 & 3096.22 & 1548.11 & 277.45 & 27.97 & 0.8120 &  19.25 & 6.21 \\
                JSI-GAN \cite{Kim_Oh_Kim_2020} & 3030.00 & 3188.78 & 1594.39 & 912.27 & 28.98 & 0.8682 & 13.72 & 6.78\\
                \rowcolor{gray!10}
                {FDAN}~\cite{xu2023fdan} & {142.25} & {116.46} & {58.83} & {50.19} & {31.60} & {0.9018} & {14.66} & {7.34} \\
                \hhline{-||----||----}
                \textbf{IRNet-5 (64c)} & 468.19 & 503.01 & 251.35 & 64.91 & {\color{red} \textbf{33.74}} & {\color{blue} \textbf{0.9150}} & {\color{blue} \textbf{10.94}} & {\color{blue} \textbf{7.55}} \\
        \hline
         
            
    \end{NiceTabular}
    \label{tab:SR-ITMtable}
    \vspace{-3mm}
\end{table*}

\noindent
\textbf{Visual quality}. In Figure~\ref{com_SRITM}, we qualitatively compare the visual results of different methods on the HDRTV1K test set~\cite{chen2021new} modified for joint SR-ITM (please refer to \S\ref{dataset}). One can see that all these methods obtain promising visual results on the presented scenes. The method of ``HDRTVNet+RFDN'' produces blurry edges around the lighting area. Besides, the images output by ``HDRTVNet+RFDN'', ``RFDN+HDRTVNet'', JSI-GAN \cite{Kim_Oh_Kim_2020} and FDAN \cite{xu2023fdan} suffer from the color shift problem to some extent. By fully exploiting multi-layer features for fine-grained image reconstruction, our IRNet-5 (64c) not only accurately restores the image colors, but also well increases the image details during the SR process. These results validate that, though being lightweight with the fewest parameter amounts and computational costs, the proposed IRNet is very efficient on the joint SR-ITM task.

\noindent
\textbf{Running speed}. The comparison running speeds on the downsampled images ($960\times540\times3$) are summarized in Table \ref{tab:SR-ITMtable}. One can see that our IRNet is faster than other comparison methods. Note that when comparing with ``RFDN+HDRTVNet'', our IRNet-5 achieves comparable performance with only 4.08\% of its running time. These results validate the efficiency of our IRNet on joint SR-ITM.

\subsection{Ablation Study}
\label{ablation study}

To study in detail the working mechanism of our IRNet, we present comprehensive ablation experiments of our IRNet on ITM. Specifically, we assess:
1) how to extract the intermediate feature $F_1$ in our IRB?
2) how does the number of IRB blocks affect our IRNet?
3) how does the channel dimension $C$ in IRB influence our IRNet? 
4) how does the CCA layer boost our IRNet?
All variants of our IRNet are trained on the training set of HDRTV1K~\cite{chen2021new}. Their performance on ITM tasks are evaluated on the test set of HDRTV1K~\cite{chen2021new}, ITM-4K and the test set in \cite{9008274}, while the performance on joint SR-ITM tasks are evaluated on the test set of HDRTV1K~\cite{chen2021new}.

\begin{table*}[!t]\footnotesize
    \centering
    \caption{ \textbf{Results of our IRNet-2 with different variants of IRB for ITM} on the test set in HDRTV1K \cite{chen2021new}, ITM-4K and the test set in \cite{9008274}. The parameter amount of our IRNet replacing IRB by RB with LeakyReLU is 158.92K, while our IRNet has 134.73K parameters.}
    \vspace{-1mm}
    \renewcommand{\arraystretch}{1.2}
    \setlength{\tabcolsep}{1.5mm}
    
    \begin{NiceTabular}{c||c||c|c|c|c}
        \Xhline{1.4pt}
            Test Dataset & Variant of IRB  & PSNR(dB)$\uparrow$ & SSIM$\uparrow$ & LPIPS(\%)$\downarrow$ & HDR-VDP3$\uparrow$\\
        \hhline{=||=|=|=|=|=}
        \multirow{5}{*}{test set in HDRTV1K\cite{chen2021new}} & RB with LeakyReLU & 36.96 & 0.9728 & 2.90 & 8.54 \\
        \rowcolor[gray]{0.9}
        \cellcolor{white} & IRB & \textbf{38.20} & \textbf{0.9737} & 2.68 & 8.58 \\
        & IRB w/o $F_1$ & 37.91 & 0.9731 & 2.69 & \textbf{8.59} \\
        & Take $F_{in}$ as $F_1$ & 37.60 & 0.9735 & 2.66 & 8.58 \\
        & Take $F_2$ as $F_1$ & 37.87 & 0.9733 & \textbf{2.64} & 8.58 \\
        \hhline{=||=||=|=|=|=}
        \multirow{5}{*}{{TM-4K}} & {RB with LeakyReLU} & 34.80 & 0.9459 & 5.41 & {8.21}\\
        \rowcolor[gray]{0.9}
        \cellcolor{white} & {IRB} & \textbf{35.24} & {0.9459} & {5.48} & \textbf{8.23}\\
        & {IRB w/o $F_1$} & 34.93 & 0.9454 & 5.38 & 8.21 \\
        & { Take $F_{in}$ as $F_1$} & 34.91 & 0.9452 & 5.33 & 8.21\\
        & { Take $F_2$ as $F_1$} & 35.07 & \textbf{0.9461} & \textbf{5.32} & { 8.22}\\
        \hhline{=||=||=|=|=|=}
        \multirow{5}{*}{{ test set in} \cite{9008274}} & { RB with LeakyReLU} & 32.67 & 0.9592 & 3.15 & 8.13\\
        \rowcolor[gray]{0.9}
        \cellcolor{white} & { IRB} &  \textbf{33.85} & 0.9609 & 2.90 & 8.20 \\
        & {IRB w/o $F_1$} & 32.93 & 0.9606 & 3.03 & 8.21 \\
        & { Take $F_{in}$ as $F_1$} & 33.10 & 0.9606 & 3.03 & \textbf{8.22}\\
        & {Take $F_2$ as $F_1$} & 33.65 &  \textbf{0.9612} & \textbf{2.80} & 8.20\\
        \hline
    \end{NiceTabular}

    \label{tab:IRB importance}
    \vspace{-4mm}
\end{table*}

\noindent
\textbf{1) How to extract the intermediate feature $F_1$ in our IRB}? The IRB in our IRNet is modified from the residual block (RB). To validate the effectiveness of our IRB, we first evaluate our IRNet by replacing the IRB blocks by the RB blocks (using LeakyReLU instead of ReLU for fair comparison). The results listed in the first two rows of Table \ref{tab:IRB importance} show that our IRNet with the IRB block achieves much better performance than our IRNet with the original RB block.

Besides, we design several variants of our IRB block (``IRB'') and study how they influence our IRNet on ITM. We first remove the intermediate feature $F_1$ to verify its importance in our IRB, which is denoted as ``IRB w/o $F_1$''. Then we study where to extract the intermediate feature $F_1$, which can be put before the first convolution layer (take $F_1$ as $F_{in}$), after the activation layer (our IRB), before the addition operation (take $F_1$ as $F_2$). The results on three test sets are summarized in Table \ref{tab:IRB importance}. One can see that our IRNet with the original IRB achieves the best PSNR results on all test sets. By removing the feature $F_1$, the variant of our IRNet achieves a clear drop on PSNR results, while the SSIM and LPIPS results drop on some test sets. If we use the input feature $F_{in}$ of IRB or the feature after the second convolution layer $F_2$ as the intermediate feature $F_{in}$, the variants of our IRNet achieve lower PSNR results, with slight difference on SSIM, LPIPS and HDR-VDP3. All these results validate the effectiveness of utilizing the intermediate feature $F_1$ after LeakyReLU as the intermediate feature of our IRB for promising ITM.

\begin{table*}[t]\footnotesize
    \centering
    \caption{\textbf{Results of our IRNet with different number of IRB blocks for ITM} on the test set in HDRTV1K \cite{chen2021new}, ITM-4K, and the test set in \cite{9008274}.}
    \vspace{-1mm}
    \renewcommand{\arraystretch}{1.2}
    \setlength{\tabcolsep}{1.3mm}
    \begin{NiceTabular}{c||c||c||c|c|c|c}
        \Xhline{1.4pt}
            Test Dataset & \#Block & Params(K)$\downarrow$ & PSNR(dB)$\uparrow$ & SSIM$\uparrow$ & LPIPS(\%)$\downarrow$ & HDR-VDP3$\uparrow$\\
        \hhline{=||=||=||=|=|=|=}
            & 1 & 86.86 & 37.76 & 0.9735 & 2.69 & 8.56\\
            \rowcolor[gray]{0.9}
            \cellcolor{white} & 2 & 134.73 & \textbf{38.20} & \textbf{0.9737} & 2.68 & \textbf{8.58} \\
            & 3 & 182.61 & 37.79 & \textbf{0.9737} & 2.64 & 8.56 \\
            \multirow{-4}{*}{test set in HDRTV1K\cite{chen2021new}} & 4 & 230.48 & 37.56 & 0.9736 & \textbf{2.63} & 8.57\\
        \hhline{=||=||=||=|=|=|=}
            & {1} & {86.86} & {34.85} & {\textbf{0.9460}} & {\textbf{5.35}} & 8.20\\
            \rowcolor[gray]{0.9}
            \cellcolor{white} & {2} & {134.73} & {\textbf{35.24}} & {0.9459} & {5.48} & {\textbf{8.23}} \\
            & {3} & {182.61} & {35.05} & {0.9452} & {5.38} & {\textbf{8.23}}\\
            \multirow{-4}{*}{{ ITM-4K}} & {4} & {230.48} & {34.96} & {0.9452} & {5.39} & {8.22}\\
        \hhline{=||=||=||=|=|=|=}
            & {1} & {86.86} & {33.40} & {0.9605} & {2.99} & {8.18}\\
            \rowcolor[gray]{0.9}
            \cellcolor{white} & {2} & {134.73} & {\textbf{33.85}} & {\textbf{0.9609}} & {\textbf{2.90}} & {\textbf{8.20}} \\
            & {3} & {182.61} & {33.08} & {\textbf{0.9609}} & {2.99} & {8.19}\\
            \multirow{-4}{*}{{test set in} \cite{9008274}} & {4} & {230.48} & {33.47} & {0.9608} & {2.94} & {8.18}\\
        \hline
    \end{NiceTabular}
    \label{tab:IRBs}
    \vspace{-3mm}
\end{table*}

\begin{table*}[t]\footnotesize
    \centering
    \caption{\textbf{Results of our IRNet with different number of IRB blocks for joint SR-ITM} on the test set in \cite{chen2021new}.}
    \vspace{-1mm}
    \renewcommand{\arraystretch}{1.2}
    \setlength{\tabcolsep}{4.2mm}
    \begin{NiceTabular}{c||c||c|c|c|c}
        \Xhline{1.4pt}
            \#Block & Params(K)$\downarrow$ & PSNR(dB)$\uparrow$ & SSIM$\uparrow$ & LPIPS(\%)$\downarrow$ & HDR-VDP3$\uparrow$\\
        \hhline{=||=||=||=|=|=}
            1 & 276.69 & 33.26 & 0.9129 & 11.27 & 7.50 \\
            2 & 324.56 & 33.33 & 0.9127 & 11.40 & 7.53\\
            3 & 372.44 & 33.42 & 0.9136 & 11.34 & 7.53\\
            4 & 420.32 & 33.58 & 0.9145 & 11.00 & 7.53\\
            \rowcolor[gray]{0.9}
            5 & 468.19 & \textbf{33.74} & \textbf{0.9150} & 10.94 & 7.55\\
            6 & 516.07 & 33.68 & \textbf{0.9150} & \textbf{10.81} & \textbf{7.56}\\
        \hline
    \end{NiceTabular}
    
    \label{tab:SRITM}
    \vspace{-3mm}
\end{table*}

\begin{table*}[t]\footnotesize
    \centering
    \caption{\textbf{Results of our IRB-1 with different channel dimensions} for ITM on the test set in HDRTV1K \cite{chen2021new}, ITM-4K and the test set in \cite{9008274}.}
    \renewcommand{\arraystretch}{1.2}
    \setlength{\tabcolsep}{1.2mm}
    \begin{NiceTabular}{c||c||c||c|c|c|c}
        \Xhline{1.4pt}
            Test Dataset & Channel & Params(K)$\downarrow$ & PSNR(dB)$\uparrow$ & SSIM$\uparrow$ & LPIPS(\%)$\downarrow$ & HDR-VDP3$\uparrow$\\
        \hhline{=||=||=||=|=|=|=}
            & 32 & 22.31 & 36.89 & 0.9723 & 2.79 & 8.50 \\ 
            \rowcolor[gray]{0.9}
            \cellcolor{white} & 48 & 49.30 & \textbf{37.82} & 0.9730 & \textbf{2.62} & \textbf{8.54}\\ 
            \multirow{-3}{*}{test set in HDRTV1K\cite{chen2021new}} & 64 & 86.86 & 37.76 & \textbf{0.9735} & 2.69 & \textbf{8.54}\\ 
        \hhline{=||=||=||=|=|=|=}
            & {32} & {22.31} & {34.74} & {0.9442} & \textbf{5.12} & {8.20}\\ 
            \rowcolor[gray]{0.9}
            \cellcolor{white} & {48} & {49.30} & {34.81} & {0.9452} & {5.17} & {\textbf{8.21}}\\ 
            \multirow{-3}{*}{{ITM-4K}} & {64} & 86.86 & {\textbf{34.85}} & \textbf{0.9460} & 5.35 & 8.20\\ 
        \hhline{=||=||=||=|=|=|=}
            & {32} & {22.31} & {32.69} & {0.9586} & {3.05} & {8.15}\\ 
            \rowcolor[gray]{0.9}
            \cellcolor{white} & {48} & {49.30} & {\textbf{33.67}} & {0.9595} & {\textbf{2.80}} & {8.09}\\ 
            \multirow{-3}{*}{{test set in \cite{9008274}}} & {64} & {86.86} & {33.40} & {\textbf{0.9605}} & {2.99} & {\textbf{8.18}}\\ 
        \hline
    \end{NiceTabular}
    
    \label{tab:IRB-1}
    \vspace{-4mm}
\end{table*}

\noindent
\textbf{2) How does the number of IRB blocks affect our IRNet}? In our IRNet, we use two IRB blocks for ITM and five IRB blocks for joint SR-ITM. Here, we vary the number of IRB blocks to study how it influences our IRNet. The results are listed in Tables \ref{tab:IRBs} and \ref{tab:SRITM}, respectively. It can be seen that our IRNet achieves promising performance with 1$\sim$4 IRB blocks on all metrics. Our IRNet with two IRB blocks achieves the best PSNR and HDR-VDP3 results among all choices. Similarly, our IRNet with five IRB blocks achieves the best PSNR and SSIM results on joint SR-ITM, while that with six IRB blocks achieves the best LPIPS and HDR-VDP3 results. To reduce the parameter amounts, we use two and five IRB blocks in our IRNet for ITM and joint SR-ITM, respectively.

\noindent
\textbf{3) How does the channel dimension $C$ in IRB influence our IRNet}? To answer this question, we perform experiments on our IRNet with different number of channels in the IRB block. The results of our IRNet-1 and IRNet-2 on ITM and those of our IRNet-5 on joint SR-ITM are shown in the Table \ref{tab:IRB-1}, Table \ref{tab:IRB-2} and Table \ref{tab:SRITM-c}, respectively.

For ITM, our IRNet-1 using one IRB achieves the best PSNR, LPIPS and HDR-VDP3 results on the test set of \cite{chen2021new} when $C=48$ and with 49.30K parameters, while our IRNet-2 using two IRBs achieves the best PSNR and SSIM results among all choices when $C=64$ and with 134.73K parameters. For joint SR-ITM, our IRNet-5 using five IRBs achieves the best PSNR results when $C=64$ and with 468.19K parameters. Our IRNet-5 with $C=96$ achieves better SSIM, LPIPS, and HDR-VDP3 results, but suffers from a huge growth of parameter amounts. Thus, we set $C=48$ and $C=64$ in our IRNet-1 and IRNet-2, respectively for ITM, and $C=64$ in our IRNet-5 for joint SR-ITM.

\begin{table*}[tp]\footnotesize
    \centering
    \caption{\textbf{Results of our IRB-2 with different channel dimensions} for ITM on the test set in HDRTV1K \cite{chen2021new}, ITM-4K and the test set in \cite{9008274}.}
    \renewcommand{\arraystretch}{1.2}
    \setlength{\tabcolsep}{1.2mm}
    \begin{NiceTabular}{c||c||c||c|c|c|c}
        \Xhline{1.4pt}
            Test Dataset & Channel & Params(K)$\downarrow$ & PSNR(dB)$\uparrow$ & SSIM$\uparrow$ & LPIPS(\%)$\downarrow$ & HDR-VDP3$\uparrow$ \\
        \hhline{=||=||=||=|=|=|=}
            & 32 & 34.34 & 37.34 & 0.9725 & 2.83 & 8.52\\ 
            & 48 & 76.28 & 38.11 & \textbf{0.9737} & \textbf{2.60} & 8.57\\ 
            \rowcolor[gray]{0.9}
            \cellcolor{white} & 64 & 134.73 & \textbf{38.20} & \textbf{0.9737} & 2.68 & \textbf{8.58} \\ 
            \multirow{-4}{*}{test set in HDRTV1K\cite{chen2021new}} & 96 & 301.17 & 38.06 & \textbf{0.9737} & 2.67 & \textbf{8.58}\\
        \hhline{=||=||=||=|=|=|=}
            & {32} & {34.34} & {34.95} & {0.9441} & {\textbf{5.25}} & {8.20}\\ 
            & {48} & {76.28} & {34.97} & {0.9437} & {5.26} & {8.21}\\ 
            \rowcolor[gray]{0.9}
            \cellcolor{white} & {64} & {134.73} & {\textbf{35.24}} & {\textbf{0.9459}} & {5.48} & {\textbf{8.23}} \\ 
            \multirow{-4}{*}{{ITM-4K}} & {96} & {301.17} & {35.11} & {0.9453} & {5.41} & {8.22}\\
        \hhline{=||=||=||=|=|=|=}
            & {32} & {34.34} & {33.17} & {0.9596} & {3.05} & {8.17}\\ 
            & {48} & {76.28} & {33.21} & {0.9599} & {3.01} & {8.17}\\ 
            \rowcolor[gray]{0.9}
            \cellcolor{white} & {64} & {134.73} & {\textbf{33.85}} & {\textbf{0.9609}} & {2.90} & {8.20} \\ 
            \multirow{-4}{*}{{test set in \cite{9008274}}} & {96} & {301.17} & {33.42} & {\textbf{0.9609}} & {\textbf{2.86}} & {\textbf{8.22}}\\
        \hline
    \end{NiceTabular}
    
    \label{tab:IRB-2}
    \vspace{-4mm}
\end{table*}

\begin{table*}[pt]\footnotesize
    \centering
    \caption{\textbf{Results of our IRB-5 with different channel dimensions} for joint SR-ITM on the test set \cite{chen2021new}.}
    \vspace{-1mm}
    \renewcommand{\arraystretch}{1.2}
    \setlength{\tabcolsep}{4.1mm}
    \begin{NiceTabular}{c||c||c|c|c|c}
        \Xhline{1.4pt}
            Channel & Params(K)$\downarrow$ & PSNR(dB)$\uparrow$ & SSIM$\uparrow$ & LPIPS(\%)$\downarrow$ & HDR-VDP3$\uparrow$\\
        \hhline{=||=||=|=|=|=}
            32 & 119.29 & 33.18 & 0.9114 & 11.68 & 7.50\\
            48 & 265.04 & 33.60 & 0.9133 & 11.08 & 7.54\\
            \rowcolor[gray]{0.9}
            64 & 468.19 & \textbf{33.74} & \textbf{0.9150} & 10.94 & 7.55\\
            96 & 1046.73 & 33.62 & \textbf{0.9150} & \textbf{10.46} & \textbf{7.57}\\
        \hline
    \end{NiceTabular}
    
    \label{tab:SRITM-c}
    \vspace{-4mm}
\end{table*}

\begin{table*}[p!t]\footnotesize
    \centering
    \caption{\textbf{Results of our IRNet-2 with different settings of CCA layers~\cite{CCA}} on the test set in HDRTV1K \cite{chen2021new}, ITM-4K and the test set in \cite{9008274}.}
    \vspace{-1mm}
    \renewcommand{\arraystretch}{1.2}
    \setlength{\tabcolsep}{1.4mm}
    \begin{NiceTabular}{c||c||c|c|c|c}
        \Xhline{1.6pt}
            Test Dataset & Method (IRNet) & PSNR(dB)$\uparrow$ & SSIM$\uparrow$ & LPIPS(\%)$\downarrow$ & HDR-VDP3$\uparrow$ \\
        \hhline{=||=||=|=|=|=}
            \rowcolor[gray]{0.9}
        \cellcolor{white} & w/ CCA layers & \textbf{38.20} & \textbf{0.9737} & \textbf{2.68} & \textbf{8.58} \\
            \multirow{-2}{*}{test set in HDRTV1K\cite{chen2021new}} & w/o 1-st CCA Layer & 37.73 & 0.9733 & 2.69 & \textbf{8.58}\\
        \hhline{=||=||=|=|=|=}
            \rowcolor[gray]{0.9}
        \cellcolor{white} & {w/ CCA layers} & {\textbf{35.24}} & {\textbf{0.9459}} & {5.48} & {\textbf{8.23}} \\
            \multirow{-2}{*}{{ITM-4K}} & {w/o 1-st CCA Layer} & {34.97} & {0.9733} & {\textbf{5.45}} & {8.22}\\
        \hhline{=||=||=|=|=|=}
            \rowcolor[gray]{0.9}
        \cellcolor{white} & {w/ CCA layers} & {\textbf{33.85}} & {\textbf{0.9609}} & {\textbf{2.90}} & {\textbf{8.20}} \\
            \multirow{-2}{*}{{test set in} \cite{9008274}} & {w/o 1-st CCA Layer} & {33.15} & {0.9608} & {2.97} & {8.17}\\
        \hline
    \end{NiceTabular}
    \label{tab:CCA}
    \vspace{-4mm}
\end{table*}


\noindent
\textbf{4) How does the CCA layer~\cite{CCA} boost our IRNet}? Our IRNet uses one CCA layer~\cite{CCA} after each IRB block to refine the feature maps. We remove the first CCA layer between two IRB blocks in our IRNet-2. The results on ITM are shown in Table \ref{tab:CCA}. One can see that our IRNet-2 without the first CCA layer~\cite{CCA} suffers from a clear performance drop on PSNR among all test sets. This demonstrates that the CCA layer~\cite{CCA} is important to our IRNet-2 on ITM.

\begin{figure*}[pt]
    \centering \includegraphics[width=13cm]{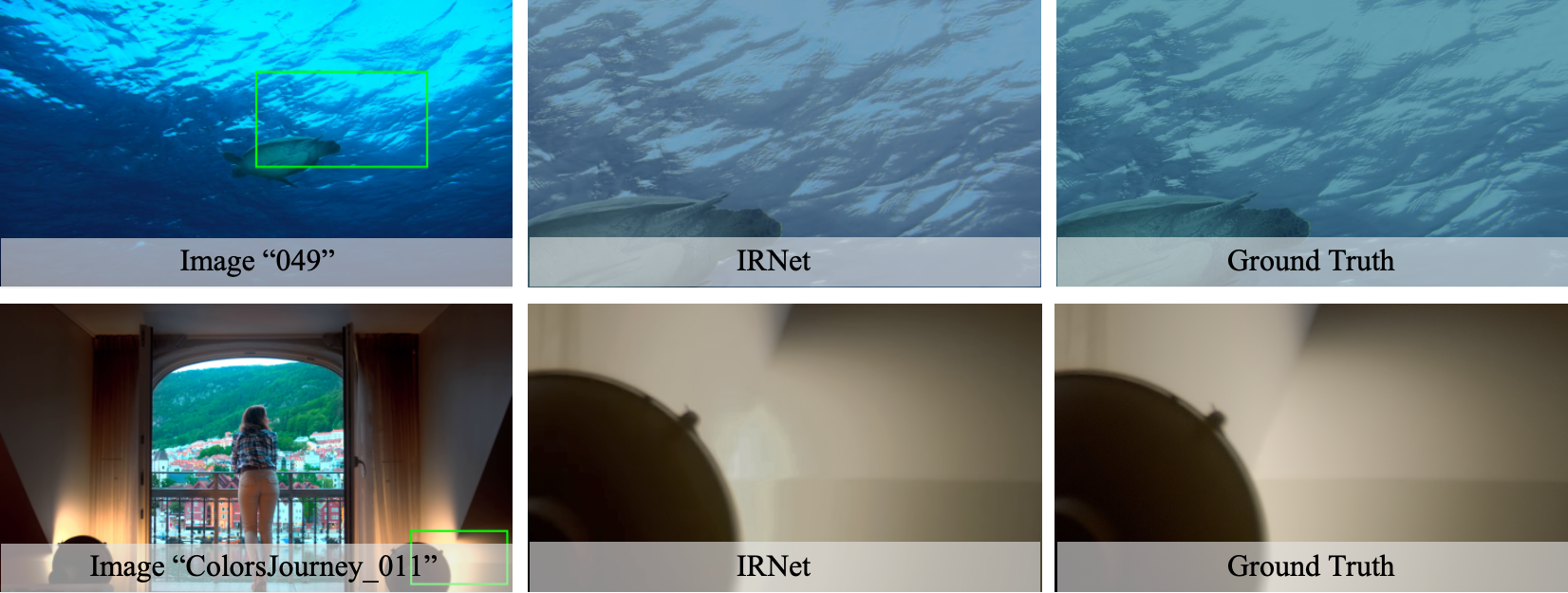}
    \vspace{-3mm}
    \caption{{\bfseries Failure case of our IRNet on the HDRTV1K test set~\cite{chen2021new} (1-st row) and our ITM-4K test dataset (2-nd row).
    }}\label{fig:failure_case}
\vspace{-3mm}
\end{figure*}

\subsection{Failure Case} \label{sec:failurecase}
To understand the limitation of the proposed IRNet, we provide two failure cases of our IRNet in Figure \ref{fig:failure_case}. 
For the 1-st row, compared with ground truth, the output image of our IRNet suffers from color bias: the vivid blue in the input SDR image is inaccurately restored to grayish blue, while the corresponding HDR image exhibits a light green-blue tone.
Besides, in the 2-nd row, the image generated by our IRNet has an unnatural region near the light source: the brightness information extracted by our IRNet is insufficient for recover consistent brightness transition area from light to dark.

\section{Conclusion}
\label{conclusion} 
In this paper, we developed a lightweight and efficient inverse tone mapping (ITM) network. The proposed Improved Residual Network (IRNet) mainly consists of Improved Residual Blocks (IRB) modified from the popular residual block and Contrast-aware Channel Attention (CCA) layers. The proposed IRB block is able to fuse multi-layer features extracted by different convolution layers for fine-grained ITM. We also collected a new ITM-4K test set containing 160 versatile 4K-resolution SDR images. Experiments on three benchmark datasets demonstrated that, our IRNet outperforms the state-of-the-art methods on the ITM task with only $\sim$0.13M parameters and $\sim$0.22$\times10^4$G FLOPs per 4K image. Further experiments on the joint SR-ITM task also showed the advantages of our IRNet over the comparison methods on the objective metrics, the computational efficiency, and most importantly, the image quality such as color depth restoration.

Although our IRNet has achieved promising results on ITM and joint SR-ITM, it sometimes produces subtle color shift and unnatural brightness transition upon challenging cases. As a potential future work, we will address color shift by employing a color correction module~\cite{corr2023} as a post-processing step. Besides, the issue of the unnatural brightness transition could be possibly alleviated by utilizing stronger feature fusion techniques~\cite{cui2021tf,liu2023tfp,2019rgb_d_fusion,liu2021densernet,2021_sg_net}.

\section*{Declarations}
\begin{itemize}
\item \textbf{Funding} This research is supported by the National Natural Science Foundation of China (No. 62002176, 62176068, 12101334, and 62171309), CAAI-Huawei MindSpore Open Fund, the Natural Science Foundation of Tianjin (No. 21JCQNJC00030), the Open Research Fund (No. B10120210117-OF03) from the Guangdong Provincial Key Laboratory of Big Data Computing, The Chinese University of Hong Kong, Shenzhen, and the Fundamental Research Funds for the Central Universities.

\item \textbf{Conflict of interest} The authors declare no conflict of interest.

\item \textbf{Availability of data and materials} The authors declare that all the data associated with the manuscript is mentioned in the manuscript. The HDRTV1K dataset ~\cite{chen2021new} and the test set in~\cite{9008274} are publicly released, and are available from the corresponding GitHub homepage. The ITM-4K dataset we collected will be publicly available at \url{https://github.com/ThisisVikki/ITM-baseline}.

\end{itemize}

\bibliography{reference}

\begin{thebibliography}{73}
\providecommand{\natexlab}[1]{#1}
\providecommand{\url}[1]{{#1}}
\providecommand{\urlprefix}{URL }
\providecommand{\doi}[1]{\url{https://doi.org/#1}}
\providecommand{\eprint}[2][]{\url{#2}}
 \bibcommenthead

\bibitem[{Afifi et~al(2021)Afifi, Derpanis, Ommer, and
  Brown}]{afifi2021learning}
Afifi M, Derpanis KG, Ommer B, et~al (2021) Learning multi-scale photo exposure
  correction. In: IEEE Conf. Comput. Vis. Pattern Recog., pp 9157--9167

\bibitem[{Aky\"{u}z et~al(2007)Aky\"{u}z, Fleming, Riecke, Reinhard, and
  B\"{u}lthoff}]{Aky2007}
Aky\"{u}z AO, Fleming R, Riecke BE, et~al (2007) Do hdr displays support ldr
  content? a psychophysical evaluation. In: ACM SIGGRAPH. Association for
  Computing Machinery, New York, NY, USA, SIGGRAPH '07, p 38–es,
  \doi{10.1145/1275808.1276425},
  \urlprefix\url{https://doi.org/10.1145/1275808.1276425}

\bibitem[{Banterle et~al(2006)Banterle, Ledda, Debattista, and
  Chalmers}]{banterle2006inverse}
Banterle F, Ledda P, Debattista K, et~al (2006) Inverse tone mapping. In:
  Proceedings of the 4th international conference on Computer graphics and
  interactive techniques in Australasia and Southeast Asia, pp 349--356

\bibitem[{Banterle et~al(2008)Banterle, Ledda, Debattista, and
  Chalmers}]{Banterle08}
Banterle F, Ledda P, Debattista K, et~al (2008) Expanding low dynamic range
  videos for high dynamic range applications. In: Proceedings of the 24th
  Spring Conference on Computer Graphics. Association for Computing Machinery,
  New York, NY, USA, SCCG '08, p 33–41, \doi{10.1145/1921264.1921275},
  \urlprefix\url{https://doi.org/10.1145/1921264.1921275}

\bibitem[{Boitard et~al(2018)Boitard, Pourazad, and
  Nasiopoulos}]{Boitard2018ColorGamut}
Boitard R, Pourazad MT, Nasiopoulos P (2018) Compression efficiency of high
  dynamic range and wide color gamut pixel’s representation. IEEE
  Transactions on Broadcasting 64(1):1--10. \doi{10.1109/TBC.2017.2781120}

\bibitem[{Burt and Adelson(1987)}]{burt1987laplacian}
Burt PJ, Adelson EH (1987) The laplacian pyramid as a compact image code. In:
  Readings in Computer Vision. Elsevier, p 671--679

\bibitem[{Chen et~al(2016)Chen, Adams, Wadhwa, and Hasinoff}]{chen2016}
Chen J, Adams A, Wadhwa N, et~al (2016) Bilateral guided upsampling. ACM Trans
  Graph 35(6). \doi{10.1145/2980179.2982423},
  \urlprefix\url{https://doi.org/10.1145/2980179.2982423}

\bibitem[{Chen et~al(2022)Chen, Yang, Chan, Li, Hou, and Chau}]{Chen2022HDRre}
Chen J, Yang Z, Chan TN, et~al (2022) Attention-guided progressive neural
  texture fusion for high dynamic range image restoration. IEEE Trans Image
  Process 31:2661--2672. \doi{10.1109/TIP.2022.3160070}

\bibitem[{Chen et~al(2021{\natexlab{a}})Chen, Liu, Zhang, Qiao, and
  Dong}]{chen2021hdrunet}
Chen X, Liu Y, Zhang Z, et~al (2021{\natexlab{a}}) Hdrunet: Single image hdr
  reconstruction with denoising and dequantization. In: IEEE Conf. Comput. Vis.
  Pattern Recog., pp 354--363

\bibitem[{Chen et~al(2021{\natexlab{b}})Chen, Zhang, Ren, Tian, Qiao, and
  Dong}]{chen2021new}
Chen X, Zhang Z, Ren JS, et~al (2021{\natexlab{b}}) A new journey from sdrtv to
  hdrtv. In: Int. Conf. Comput. Vis., pp 4500--4509

\bibitem[{Cheng et~al(2022)Cheng, Wang, Li, Song, Chen, and
  Xiong}]{chengHDRTVRe2022}
Cheng Z, Wang T, Li Y, et~al (2022) Towards real-world hdrtv reconstruction: A
  data synthesis-based approach. In: Eur. Conf. Comput. Vis. Springer Nature
  Switzerland, Cham, pp 199--216

\bibitem[{Cui et~al(2021)Cui, Yan, Cao, and Liu}]{cui2021tf}
Cui Y, Yan L, Cao Z, et~al (2021) Tf-blender: Temporal feature blender for
  video object detection. In: Int. Conf. Comput. Vis., pp 8138--8147

\bibitem[{Dong et~al(2016)Dong, Loy, and Tang}]{dong2016accelerating}
Dong C, Loy CC, Tang X (2016) Accelerating the super-resolution convolutional
  neural network. In: Eur. Conf. Comput. Vis., Springer, pp 391--407

\bibitem[{Eilertsen et~al(2017)Eilertsen, Kronander, Denes, Mantiuk, and
  Unger}]{eilertsen2017hdr}
Eilertsen G, Kronander J, Denes G, et~al (2017) Hdr image reconstruction from a
  single exposure using deep cnns. ACM Trans Graph 36(6):1--15

\bibitem[{Endo et~al(2017)Endo, Kanamori, and Mitani}]{endo}
Endo Y, Kanamori Y, Mitani J (2017) Deep reverse tone mapping. ACM Trans Graph
  36:1--10. \doi{10.1145/3130800.3130834}

\bibitem[{Gao et~al(2021)Gao, Cheng, Zhao, Zhang, Yang, and Torr}]{res2net}
Gao SH, Cheng MM, Zhao K, et~al (2021) Res2net: A new multi-scale backbone
  architecture. IEEE Transactions on Pattern Analysis and Machine Intelligence
  43(2):652--662. \doi{10.1109/TPAMI.2019.2938758}

\bibitem[{Gharbi et~al(2017)Gharbi, Chen, Barron, Hasinoff, and
  Durand}]{gharbi2017deep}
Gharbi M, Chen J, Barron JT, et~al (2017) Deep bilateral learning for real-time
  image enhancement. ACM Trans Graph 36(4):1--12

\bibitem[{Goodfellow et~al(2014)Goodfellow, Pouget-Abadie, Mirza, Xu,
  Warde-Farley, Ozair, Courville, and Bengio}]{gan}
Goodfellow I, Pouget-Abadie J, Mirza M, et~al (2014) Generative adversarial
  nets. In: Adv. Neural Inform. Process. Syst. MIT Press, Cambridge, MA, USA,
  NIPS'14, p 2672–2680

\bibitem[{He et~al(2020)He, Liu, Qiao, and Dong}]{he2020conditional}
He J, Liu Y, Qiao Y, et~al (2020) Conditional sequential modulation for
  efficient global image retouching. In: Eur. Conf. Comput. Vis., Springer, pp
  679--695

\bibitem[{He et~al(2013)He, Sun, and Tang}]{he2013guided}
He K, Sun J, Tang X (2013) Guided image filtering. IEEE Trans Pattern Anal Mach
  Intell 35(6):1397--1409. \doi{10.1109/TPAMI.2012.213}

\bibitem[{He et~al(2016)He, Zhang, Ren, and Sun}]{he2016deep}
He K, Zhang X, Ren S, et~al (2016) Deep residual learning for image
  recognition. In: IEEE Conf. Comput. Vis. Pattern Recog., pp 770--778

\bibitem[{Hou et~al(2023)Hou, Xu, Hou, Hu, Wei, and Shen}]{Hou2023FaceSR}
Hou H, Xu J, Hou Y, et~al (2023) Semi-cycled generative adversarial networks
  for real-world face super-resolution. IEEE Trans Image Process 32:1184--1199.
  \doi{10.1109/TIP.2023.3240845}

\bibitem[{Hu et~al(2022)Hu, Xu, Gu, Cheng, and Liu}]{Hu2022SR}
Hu X, Xu J, Gu S, et~al (2022) Restore globally, refine locally: A mask-guided
  scheme to accelerate super-resolution networks. In: Eur. Conf. Comput. Vis.
  Springer Nature Switzerland, Cham, pp 74--91

\bibitem[{Hu et~al(2023)Hu, Huang, Huang, Xu, and Zhou}]{hu2023dmvfn}
Hu X, Huang Z, Huang A, et~al (2023) A dynamic multi-scale voxel flow network
  for video prediction. In: IEEE Conf. Comput. Vis. Pattern Recog.

\bibitem[{Hui et~al(2019)Hui, Gao, Yang, and Wang}]{CCA}
Hui Z, Gao X, Yang Y, et~al (2019) Lightweight image super-resolution with
  information multi-distillation network. In: ACM Int. Conf. Multimedia.
  Association for Computing Machinery, New York, NY, USA, MM '19, p
  2024–2032, \doi{10.1145/3343031.3351084},
  \urlprefix\url{https://doi.org/10.1145/3343031.3351084}

\bibitem[{Isola et~al(2017)Isola, Zhu, Zhou, and Efros}]{isola2017image}
Isola P, Zhu JY, Zhou T, et~al (2017) Image-to-image translation with
  conditional adversarial networks. In: IEEE Conf. Comput. Vis. Pattern Recog.,
  pp 1125--1134

\bibitem[{{Jianyi Wang} et~al(2023){Jianyi Wang}, {Zongsheng Yue}, {Shangchen
  Zhou}, {Kelvin C. K. Chan}, and {Chen Change Loy}}]{corr2023}
{Jianyi Wang}, {Zongsheng Yue}, {Shangchen Zhou}, et~al (2023) Exploiting
  diffusion prior for real-world image super-resolution. CoRR

\bibitem[{Jo et~al(2022)Jo, Lee, Ahn, and Kang}]{9447972}
Jo SY, Lee S, Ahn N, et~al (2022) Deep arbitrary hdri: Inverse tone mapping
  with controllable exposure changes. IEEE Trans Multimedia 24:2713--2726.
  \doi{10.1109/TMM.2021.3087034}

\bibitem[{Kim et~al(2019)Kim, Oh, and Kim}]{9008274}
Kim SY, Oh J, Kim M (2019) Deep sr-itm: Joint learning of super-resolution and
  inverse tone-mapping for 4k uhd hdr applications. In: Int. Conf. Comput.
  Vis., pp 3116--3125, \doi{10.1109/ICCV.2019.00321}

\bibitem[{Kim et~al(2020)Kim, Oh, and Kim}]{Kim_Oh_Kim_2020}
Kim SY, Oh J, Kim M (2020) Jsi-gan: Gan-based joint super-resolution and
  inverse tone-mapping with pixel-wise task-specific filters for uhd hdr video.
  Association for the Advancement of Artificial Intelligence
  34(07):11287--11295. \doi{10.1609/aaai.v34i07.6789},
  \urlprefix\url{https://ojs.aaai.org/index.php/AAAI/article/view/6789}

\bibitem[{Kingma and Ba(2014)}]{kingma2014adam}
Kingma DP, Ba JL (2014) Adam: A method for stochastic optimization. arXiv
  preprint arXiv:14126980

\bibitem[{Kong et~al(2021)Kong, Zhao, Qiao, and Dong}]{classsr}
Kong X, Zhao H, Qiao Y, et~al (2021) Classsr: A general framework to accelerate
  super-resolution networks by data characteristic. In: IEEE Conf. Comput. Vis.
  Pattern Recog., pp 12016--12025

\bibitem[{Kovaleski and Oliveira(2014)}]{6915289}
Kovaleski RP, Oliveira MM (2014) High-quality reverse tone mapping for a wide
  range of exposures. In: 27th SIBGRAPI Conference on Graphics, Patterns and
  Images, pp 49--56, \doi{10.1109/SIBGRAPI.2014.29}

\bibitem[{Lecouat et~al(2022)Lecouat, Eboli, Ponce, and Mairal}]{Lecouat2022}
Lecouat B, Eboli T, Ponce J, et~al (2022) High dynamic range and
  super-resolution from raw image bursts. ACM Trans Graph 41(4).
  \doi{10.1145/3528223.3530180},
  \urlprefix\url{https://doi.org/10.1145/3528223.3530180}

\bibitem[{Li et~al(2018)Li, Fang, Mei, and Zhang}]{Li_2018_ECCV}
Li J, Fang F, Mei K, et~al (2018) Multi-scale residual network for image
  super-resolution. In: The European Conference on Computer Vision (ECCV)

\bibitem[{Liang et~al(2021)Liang, Zeng, and Zhang}]{liang2021high}
Liang J, Zeng H, Zhang L (2021) High-resolution photorealistic image
  translation in real-time: A laplacian pyramid translation network. In: IEEE
  Conf. Comput. Vis. Pattern Recog., pp 9392--9400

\bibitem[{Liang et~al(2023)Liang, Cui, Wang, Geng, Wang, and
  Liu}]{liang2023clusterformer}
Liang JC, Cui Y, Wang Q, et~al (2023) Clusterformer: Clustering as a universal
  visual learner. In: Neural Information Processing Systems (NeurIPS)

\bibitem[{Liang et~al(2018)Liang, Xu, Zhang, Cao, and Zhang}]{Liang_2018_CVPR}
Liang Z, Xu J, Zhang D, et~al (2018) A hybrid l1-l0 layer decomposition model
  for tone mapping. In: IEEE Conf. Comput. Vis. Pattern Recog.

\bibitem[{Lim et~al(2017)Lim, Son, Kim, Nah, and Lee}]{edsr}
Lim B, Son S, Kim H, et~al (2017) Enhanced deep residual networks for single
  image super-resolution. In: IEEE Conf. Comput. Vis. Pattern Recog. Worksh.,
  pp 1132--1140, \doi{10.1109/CVPRW.2017.151}

\bibitem[{Liu et~al(2021{\natexlab{a}})Liu, Cui, Tan, and Chen}]{2021_sg_net}
Liu D, Cui Y, Tan W, et~al (2021{\natexlab{a}}) Sg-net: Spatial granularity
  network for one-stage video instance segmentation. In: Proceedings of the
  IEEE/CVF Conference on Computer Vision and Pattern Recognition (CVPR), pp
  9816--9825

\bibitem[{Liu et~al(2021{\natexlab{b}})Liu, Cui, Yan, Mousas, Yang, and
  Chen}]{liu2021densernet}
Liu D, Cui Y, Yan L, et~al (2021{\natexlab{b}}) Densernet: Weakly supervised
  visual localization using multi-scale feature aggregation. In: Association
  for the Advancement of Artificial Intelligence, pp 6101--6109

\bibitem[{Liu et~al(2023)Liu, Liang, Geng, Loui, and Zhou}]{liu2023tfp}
Liu D, Liang J, Geng T, et~al (2023) Tripartite feature enhanced pyramid
  network for dense prediction. IEEE Transactions on Image Processing
  32:2678--2692. \doi{10.1109/TIP.2023.3272826}

\bibitem[{Liu et~al(2020{\natexlab{a}})Liu, Tang, and Wu}]{Liu2020}
Liu J, Tang J, Wu G (2020{\natexlab{a}}) Residual feature distillation network
  for lightweight image super-resolution. In: Eur. Conf. Comput. Vis. Springer
  International Publishing, Cham, pp 41--55

\bibitem[{Liu et~al(2020{\natexlab{b}})Liu, Zhang, Tang, Tang, and
  Wu}]{9156371}
Liu J, Zhang W, Tang Y, et~al (2020{\natexlab{b}}) Residual feature aggregation
  network for image super-resolution. In: IEEE Conf. Comput. Vis. Pattern
  Recog., pp 2356--2365, \doi{10.1109/CVPR42600.2020.00243}

\bibitem[{Liu et~al(2020{\natexlab{c}})Liu, Lai, Chen, Kao, Yang, Chuang, and
  Huang}]{liu2020single}
Liu YL, Lai WS, Chen YS, et~al (2020{\natexlab{c}}) Single-image hdr
  reconstruction by learning to reverse the camera pipeline. In: IEEE Conf.
  Comput. Vis. Pattern Recog., pp 1651--1660

\bibitem[{Loshchilov and Hutter(2016)}]{loshchilov2016sgdr}
Loshchilov I, Hutter F (2016) Sgdr: Stochastic gradient descent with warm
  restarts. arXiv preprint arXiv:160803983

\bibitem[{{Lu} et~al(2023){Lu}, {Wang}, {Ma}, {Geng}, {Chen}, {Chen}, and
  {Liu}}]{2023_transformer_flow}
{Lu} Y, {Wang} Q, {Ma} S, et~al (2023) {TransFlow: Transformer as Flow
  Learner}. arXiv e-prints arXiv:2304.11523. \doi{10.48550/arXiv.2304.11523},
  {\href{https://arxiv.org/abs/2304.11523}{{arXiv:2304.11523}}} {[cs.CV]}

\bibitem[{van~der Maaten and Hinton(2008)}]{Maaten2008VisualizingDU}
van~der Maaten L, Hinton GE (2008) Visualizing data using t-sne. Journal of
  Machine Learning Research 9:2579--2605

\bibitem[{Mantiuk et~al(2011)Mantiuk, Kim, Rempel, and
  Heidrich}]{mantiuk2011hdr}
Mantiuk R, Kim KJ, Rempel AG, et~al (2011) Hdr-vdp-2: A calibrated visual
  metric for visibility and quality predictions in all luminance conditions.
  ACM Trans Graph 30(4):1--14

\bibitem[{{Marnerides} et~al(2018){Marnerides}, {Bashford-Rogers}, {Hatchett},
  and {Debattista}}]{Marnerides2018Expandnet}
{Marnerides} D, {Bashford-Rogers} T, {Hatchett} J, et~al (2018) {ExpandNet: A
  Deep Convolutional Neural Network for High Dynamic Range Expansion from Low
  Dynamic Range Content}. arXiv e-prints arXiv:1803.02266.
  {\href{https://arxiv.org/abs/1803.02266}{{arXiv:1803.02266}}} {[cs.CV]}

\bibitem[{Masia et~al(2009)Masia, Agustin, Gutiérrez, Fleming, and
  Sorkine}]{Masia2009}
Masia B, Agustin S, Gutiérrez D, et~al (2009) Evaluation of reverse tone
  mapping through varying exposure conditions. ACM Trans Graph 28:1--8.
  \doi{10.1145/1618452.1618506}

\bibitem[{Masia et~al(2017)Masia, Serrano, and Gutierrez}]{Masia2017}
Masia B, Serrano A, Gutierrez D (2017) Dynamic range expansion based on image
  statistics. Multimedia Tools Appl 76(1):631–648

\bibitem[{{Mehta} and {Rastegari}(2021)}]{2021mobieVit}
{Mehta} S, {Rastegari} M (2021) {MobileViT: Light-weight, General-purpose, and
  Mobile-friendly Vision Transformer}. arXiv e-prints arXiv:2110.02178.
  \doi{10.48550/arXiv.2110.02178},
  {\href{https://arxiv.org/abs/2110.02178}{{arXiv:2110.02178}}} {[cs.CV]}

\bibitem[{{Mehta} and {Rastegari}(2022)}]{2022mobileVit2}
{Mehta} S, {Rastegari} M (2022) {Separable Self-attention for Mobile Vision
  Transformers}. arXiv e-prints arXiv:2206.02680.
  \doi{10.48550/arXiv.2206.02680},
  {\href{https://arxiv.org/abs/2206.02680}{{arXiv:2206.02680}}} {[cs.CV]}

\bibitem[{Meylan et~al(2007)Meylan, Daly, and Suesstrunk}]{2007Meylan}
Meylan L, Daly S, Suesstrunk S (2007) Tone mapping for high dynamic range
  displays. In: Electronic Imaging

\bibitem[{Ronneberger et~al(2015)Ronneberger, Fischer, and Brox}]{unet}
Ronneberger O, Fischer P, Brox T (2015) U-net: Convolutional networks for
  biomedical image segmentation. In: Medical Image Computing and
  Computer-Assisted Intervention, Springer, pp 234--241

\bibitem[{Santos et~al(2020)Santos, Ren, and Kalantari}]{santos2020single}
Santos MS, Ren TI, Kalantari NK (2020) Single image hdr reconstruction using a
  cnn with masked features and perceptual loss. arXiv preprint arXiv:200507335

\bibitem[{Shi et~al(2016)Shi, Caballero, Husz{\'a}r, Totz, Aitken, Bishop,
  Rueckert, and Wang}]{shi2016real}
Shi W, Caballero J, Husz{\'a}r F, et~al (2016) Real-time single image and video
  super-resolution using an efficient sub-pixel convolutional neural network.
  In: IEEE Conf. Comput. Vis. Pattern Recog., pp 1874--1883

\bibitem[{{Song} et~al(2019){Song}, {Wen}, {Fei}, and {Yu}}]{2019rgb_d_fusion}
{Song} Y, {Wen} J, {Fei} Y, et~al (2019) {Deep Robotic Prediction with
  hierarchical RGB-D Fusion}. arXiv e-prints arXiv:1909.06585.
  \doi{10.48550/arXiv.1909.06585},
  {\href{https://arxiv.org/abs/1909.06585}{{arXiv:1909.06585}}} {[cs.RO]}

\bibitem[{Tan et~al(2023)Tan, Chen, Xu, Jin, and Zhu}]{Tan2023}
Tan X, Chen H, Xu K, et~al (2023) Deep sr-hdr: Joint learning of
  super-resolution and high dynamic range imaging for dynamic scenes. IEEE
  Trans Multimedia 25:750--763. \doi{10.1109/TMM.2021.3132165}

\bibitem[{Union(2015{\natexlab{a}})}]{rec.2020}
Union IT (2015{\natexlab{a}}) {Recommendation ITU-R BT.2020-2}. Electronic
  Publication

\bibitem[{Union(2015{\natexlab{b}})}]{rec.709}
Union IT (2015{\natexlab{b}}) {Recommendation ITU-R BT.709-6}. Electronic
  Publication

\bibitem[{Wang et~al(2023)Wang, Wang, Quan, Feng, Xu, Nie, Wang, Khabsa,
  Firooz, and Liu}]{wang-etal-2023-mustie}
Wang Q, Wang J, Quan X, et~al (2023) {MUSTIE}: Multimodal structural
  transformer for web information extraction. In: Proceedings of the 61st
  Annual Meeting of the Association for Computational Linguistics (Volume 1:
  Long Papers). Association for Computational Linguistics, Toronto, Canada, pp
  2405--2420, \doi{10.18653/v1/2023.acl-long.135},
  \urlprefix\url{https://aclanthology.org/2023.acl-long.135}

\bibitem[{Wang et~al(2004)Wang, Bovik, Sheikh, and Simoncelli}]{wang2004image}
Wang Z, Bovik AC, Sheikh HR, et~al (2004) Image quality assessment: from error
  visibility to structural similarity. IEEE Trans Image Process 13(4):600--612

\bibitem[{Wang et~al(2022)Wang, Cun, Bao, Zhou, Liu, and Li}]{uformer}
Wang Z, Cun X, Bao J, et~al (2022) Uformer: A general u-shaped transformer for
  image restoration. In: IEEE Conf. Comput. Vis. Pattern Recog., pp
  17683--17693

\bibitem[{Xu et~al(2021{\natexlab{a}})Xu, Xu, Li, Wang, Sun, and
  Cheng}]{xu2021temporal}
Xu G, Xu J, Li Z, et~al (2021{\natexlab{a}}) Temporal modulation network for
  controllable space-time video super-resolution. In: IEEE Conf. Comput. Vis.
  Pattern Recog., pp 6388--6397

\bibitem[{Xu et~al(2023)Xu, chen Yang, Wang, Zhen, and Xu}]{xu2023fdan}
Xu G, chen Yang Y, Wang L, et~al (2023) Joint super-resolution and inverse
  tone-mapping: A feature decomposition aggregation network and a new
  benchmark. arXiv preprint arXiv:220703367

\bibitem[{Xu et~al(2021{\natexlab{b}})Xu, Liu, Hou, Zhen, Shao, and
  Cheng}]{Xu2021Smoothing}
Xu J, Liu ZA, Hou YK, et~al (2021{\natexlab{b}}) Pixel-level non-local image
  smoothing with objective evaluation. IEEE Trans Multimedia 23:4065--4078.
  \doi{10.1109/TMM.2020.3037535}

\bibitem[{Yan et~al(2021)Yan, Cui, Chen, and Liu}]{9414517}
Yan L, Cui Y, Chen Y, et~al (2021) Hierarchical attention fusion for
  geo-localization. In: ICASSP, pp 2220--2224,
  \doi{10.1109/ICASSP39728.2021.9414517}

\bibitem[{Yao et~al(2023)Yao, He, Li, Pan, and Xiong}]{Yao2023UHD}
Yao M, He D, Li X, et~al (2023) Bidirectional translation between uhd-hdr and
  hd-sdr videos. IEEE Trans Multimedia pp 1--15. \doi{10.1109/TMM.2023.3239656}

\bibitem[{Zeng et~al(2020)Zeng, Cai, Li, Cao, and Zhang}]{zeng2020learning}
Zeng H, Cai J, Li L, et~al (2020) Learning image-adaptive 3d lookup tables for
  high performance photo enhancement in real-time. IEEE Trans Pattern Anal Mach
  Intell

\bibitem[{Zhang et~al(2018)Zhang, Isola, Efros, Shechtman, and
  Wang}]{zhang2018unreasonable}
Zhang R, Isola P, Efros AA, et~al (2018) The unreasonable effectiveness of deep
  features as a perceptual metric. In: IEEE Conf. Comput. Vis. Pattern Recog.,
  pp 586--595

\bibitem[{Zhu et~al(2017)Zhu, Park, Isola, and Efros}]{zhu2017unpaired}
Zhu JY, Park T, Isola P, et~al (2017) Unpaired image-to-image translation using
  cycle-consistent adversarial networks. In: Int. Conf. Comput. Vis., pp
  2223--2232

\end{thebibliography}

\vfill

\end{document}